\documentclass[letterpaper, 10 pt, conference, table]{IEEEtran}
\usepackage{etex}
\IEEEoverridecommandlockouts                              %

\usepackage{etex}
\usepackage{multicol}
\usepackage[numbers,compress]{natbib}
\usepackage{amsmath,amsfonts,amssymb,amsthm}
\usepackage{float}
\usepackage{bm}
\usepackage{graphicx}
\usepackage{epstopdf}
\usepackage{epsfig}
\usepackage{xspace}
\usepackage{multirow}
\usepackage{hhline}
\usepackage[export]{adjustbox}
\usepackage{csvsimple}
\usepackage[dvipsnames]{xcolor}

\usepackage{graphicx}
\usepackage[colorinlistoftodos]{todonotes}
\newcommand{\journalVersion}[1]{}
\usepackage{xr}
\newcommand{\subparagraph}{} %
\usepackage{titlesec}
\graphicspath{{./figures/}}
\usepackage{tabularx}
\newcolumntype{Y}{>{\centering\arraybackslash}X} %

\usepackage{comment}
\usepackage{siunitx}
\usepackage{relsize}
\usepackage{ifthen}
\usepackage[colorinlistoftodos]{todonotes}

\usepackage[caption=false]{subfig}

\usepackage[vlined,ruled,linesnumbered]{algorithm2e}
\usepackage{graphics} %
\usepackage{rotating}
\usepackage{color}
\usepackage{enumerate}
\usepackage[T1]{fontenc}
\usepackage{psfrag}
\usepackage{epsfig} %
\usepackage{booktabs}
\usepackage{graphicx,url}
\usepackage{multirow}
\usepackage{array}
\usepackage{latexsym}
\usepackage{amsfonts}
\usepackage{amsmath}
\usepackage{amssymb}
\usepackage{xstring}
\usepackage[noend]{algorithmic}
\usepackage{multirow}
\usepackage{xcolor}
\usepackage{prettyref}
\usepackage{flexisym}
\usepackage{bigdelim}
\usepackage{breqn} %
\usepackage{listings}
\let\labelindent\relax
\usepackage{enumitem}
\usepackage{xspace}
\usepackage{bm}
\graphicspath{{./figures/}}
\usepackage{tikz}
\usetikzlibrary{matrix,calc}

\usepackage{mdwlist}

\let\itemize\undefined
\makecompactlist{itemize}{stditemize}

\newrefformat{prob}{Problem\,\ref{#1}}
\newrefformat{def}{Definition\,\ref{#1}}
\newrefformat{sec}{Section\,\ref{#1}}
\newrefformat{sub}{Section\,\ref{#1}}
\newrefformat{prop}{Proposition\,\ref{#1}}
\newrefformat{app}{Appendix\,\ref{#1}}
\newrefformat{alg}{Algorithm\,\ref{#1}}
\newrefformat{cor}{Corollary\,\ref{#1}}
\newrefformat{thm}{Theorem\,\ref{#1}}
\newrefformat{lem}{Lemma\,\ref{#1}}
\newrefformat{fig}{Fig.\,\ref{#1}}
\newrefformat{tab}{Table\,\ref{#1}}

\newcommand{\bdmath}{\begin{dmath}}
\newcommand{\edmath}{\end{dmath}}
\newcommand{\beq}{\begin{equation}}
\newcommand{\eeq}{\end{equation}}
\newcommand{\bdm}{\begin{displaymath}}
\newcommand{\edm}{\end{displaymath}}
\newcommand{\bea}{\begin{eqnarray}}
\newcommand{\eea}{\end{eqnarray}}
\newcommand{\beal}{\beq \begin{array}{ll}}
\newcommand{\eeal}{\end{array} \eeq}
\newcommand{\beas}{\begin{eqnarray*}}
\newcommand{\eeas}{\end{eqnarray*}}
\newcommand{\ba}{\begin{array}}
\newcommand{\ea}{\end{array}}
\newcommand{\bit}{\begin{itemize}}
\newcommand{\eit}{\end{itemize}}
\newcommand{\ben}{\begin{enumerate}}
\newcommand{\een}{\end{enumerate}}

\newcommand{\setal}{~\emph{et~al.}\xspace}
\newcommand{\eg}{\emph{e.g.,}\xspace}
\newcommand{\ie}{\emph{i.e.,}\xspace}
\newcommand{\myParagraph}[1]{{\bf #1.}\xspace}
\newcommand{\M}[1]{{\bm #1}} %
\renewcommand{\boldsymbol}[1]{{\bm #1}}

\newcommand{\hide}[1]{}

\newcommand{\hiddenText}{{\color{gray} hidden text.}}
\newcommand{\hideWithText}[1]{\hiddenText}

\DeclareMathOperator*{\argmin}{arg\,min}

\newcommand{\tran}{^{\mathsf{T}}}

\newcommand{\inv}{^{-1}}

\newcommand{\SEthree}{\ensuremath{\mathrm{SE}(3)}\xspace}

\newcommand{\ME}{\M{E}}

\newcommand{\MT}{\M{T}}
\newcommand{\MX}{\M{X}}
\newcommand{\MY}{\M{Y}}

\newcommand{\vxx}{\boldsymbol{x}}

\newcommand{\blue}[1]{{\color{blue}#1}}

\newcommand{\linkToPdf}[1]{\href{#1}{\blue{(pdf)}}}
\newcommand{\linkToPpt}[1]{\href{#1}{\blue{(ppt)}}}
\newcommand{\linkToCode}[1]{\href{#1}{\blue{(code)}}}
\newcommand{\linkToWeb}[1]{\href{#1}{\blue{(web)}}}
\newcommand{\linkToVideo}[1]{\href{#1}{\blue{(video)}}}
\newcommand{\linkToMedia}[1]{\href{#1}{\blue{(media)}}}
\newcommand{\award}[1]{\xspace} %

\newcommand{\bmat}{\left[ \begin{array}}
\newcommand{\emat}{\end{array} \right]}

\newcommand{\bal}{\begin{align}}
\newcommand{\eal}{\end{align}}

\newcommand{\hydra}{Hydra\xspace}
\newcommand{\hydraMulti}{Hydra-Multi\xspace}

\newcommand{\percFound}{\emph{\% Found}\xspace}
\newcommand{\percCorrect}{\emph{\% Correct}\xspace}
\newcommand{\positionError}{\emph{Position Error}\xspace}
\newcommand{\roomPrec}{\emph{Precision}\xspace}
\newcommand{\roomRec}{\emph{Recall}\xspace}

\usepackage[colorlinks=true, allcolors=blue, bookmarks=true]{hyperref}

\usepackage{float}
\usepackage{hyperref}
\usepackage{graphicx}
\usepackage{epstopdf}
\usepackage{epsfig}

\usepackage{xr}
\usepackage[capitalize]{cleveref}

\title{\huge{\hydraMulti: Collaborative Online Construction of\\  3D Scene Graphs with Multi-Robot Teams}}
\author{Yun Chang, Nathan Hughes, Aaron Ray, Luca Carlone
\thanks{Y. Chang, N. Hughes, A. Ray, and L. Carlone are with the Laboratory for 
Information \& Decision Systems, Massachusetts Institute of Technology, Cambridge, MA, USA, 
{\sf \{yunchang, na26933, aaronray,lcarlone\}@mit.edu}
}
\thanks{This work was
partially funded by 
ARL Distributed and Collaborative Intelligent Systems and Technology Collaborative Research Alliance (DCIST\,CRA) under agreement W911NF-17-2-0181, and by MathWorks. A.~Ray is~supported~by~a~National~Defense~Science~and~Engineering~Graduate~Fellowship.}
}

\begin{document}

\maketitle

\begin{abstract}
3D scene graphs have recently emerged as an expressive high-level map representation 
that describes a 3D environment as a layered graph where nodes represent spatial concepts at multiple levels of abstraction
(\eg objects, rooms, buildings) and edges represent relations between concepts (\eg inclusion, adjacency).
This paper describes \emph{\hydraMulti}, the first multi-robot spatial perception system capable of constructing
a multi-robot 3D scene graph online 
from sensor data collected by robots in a team.
In particular, we develop a centralized system capable of constructing a joint 3D scene graph
by taking incremental inputs from multiple robots, effectively finding the relative transforms
between the robots' frames, and incorporating loop closure detections to correctly reconcile the scene graph nodes from different robots.
{We evaluate \hydraMulti on simulated and real scenarios and
show it is able to reconstruct accurate 3D scene graphs online.
We also demonstrate \hydraMulti's capability of supporting heterogeneous teams by fusing different 
map representations built by robots with different sensor suites.} 
\end{abstract}
\section{Introduction}
\label{sec:introduction}

Multi-robot systems have become increasingly popular both in the robotics research community and in the industry at large due to their capability to sense and act over large-scale environments.
Multi-robot operation is important for applications such as factory automation, intelligent transportation, disaster response, and environmental monitoring, to name a few.

In this work, we address the problem of using a team of robots to gain situational awareness over a large environment.
In particular, we develop a system that allows robots in a team to build a high-level map representation, namely a \emph{3D scene graph}.
A 3D scene graph is a hierarchical graph where nodes represent spatial concepts  at multiple levels of abstraction (\eg objects, rooms, and buildings for indoor environments), and edges represent relations between concepts (\eg inclusion or adjacency). This representation has been recently proposed as a high-level representation of 3D environments~\cite{Armeni19iccv-3DsceneGraphs,Rosinol20rss-dynamicSceneGraphs,Rosinol21ijrr-Kimera,Hughes22rss-hydra,Wu21cvpr-SceneGraphFusion}, and has been successfully used for hierarchical planning and object search~\cite{Rosinol21ijrr-Kimera,Ravichandran22icra-RLwithSceneGraphs}.
Our goal is to collaboratively estimate a 3D scene graph of the environment that describes the spatial and semantic information of the scene the robots operate in.

3D scene graphs are particularly suitable for multi-robot operation, since they 
enable fast planning and decision-making over the large-scale environments often 
covered by multi-robot teams.
 Moreover, as shown in our experiments, their layered structure makes it easy to 
 reconcile heterogeneous map representations, \eg built by robots using different sensor suites or 
 relying on different mapping pipelines. %

\begin{figure}[!t]
\vspace{0mm}
    \centering
    \includegraphics[trim={20 20 10 20},clip, width=1.0\columnwidth]{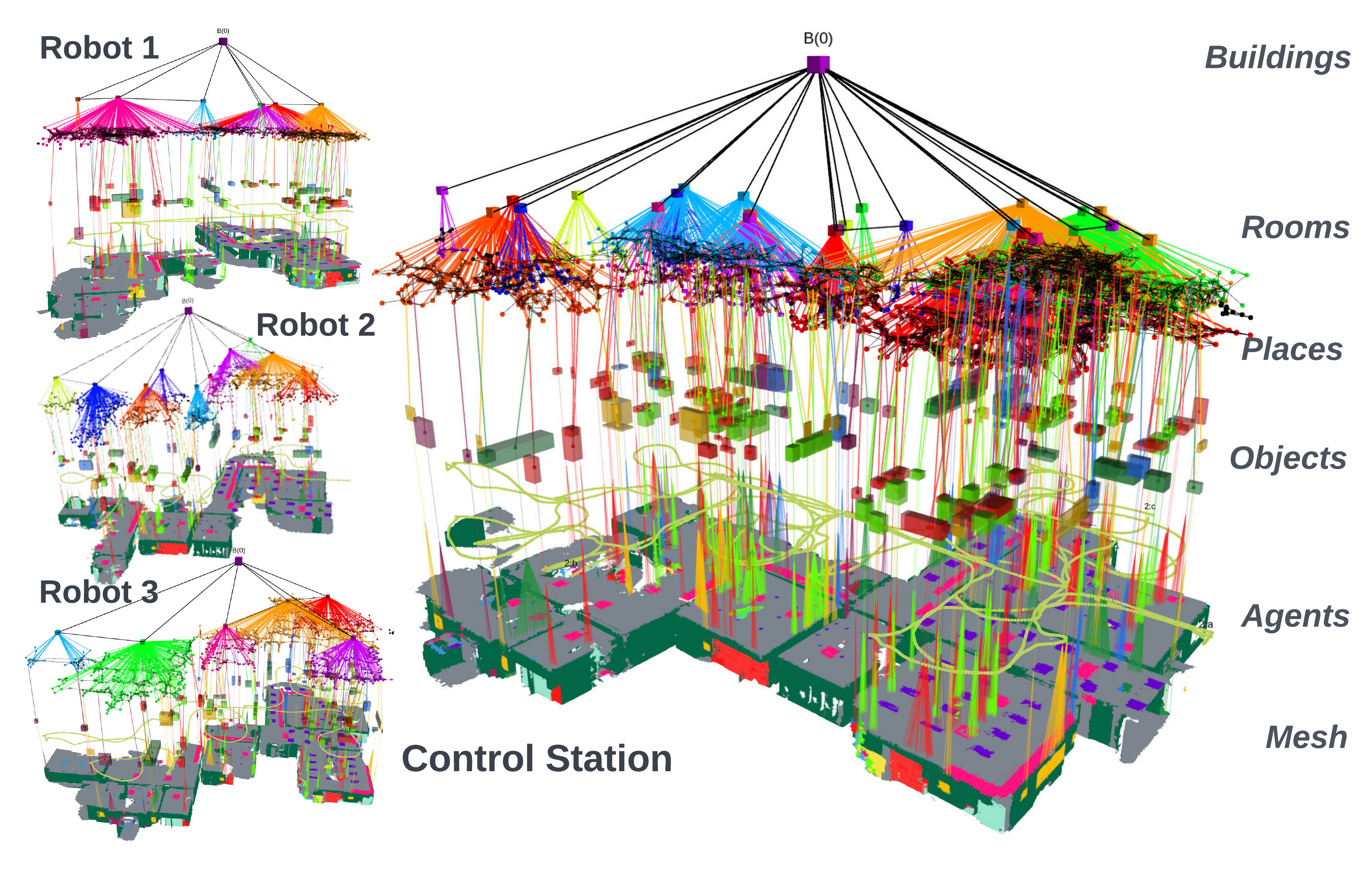}
\vspace{-8mm}
    \caption{\hydraMulti with three robots demonstrated in the uHumans2 simulated office scene.
    Each robot only explores a portion of the environment, and \hydraMulti takes in the partial single-robot scene graphs,
    and constructs a complete multi-robot 3D scene graph of the environment.}\label{fig:cover}
\end{figure}

\myParagraph{Contribution}
The main contribution of this paper is the development of \emph{\hydraMulti} (\cref{fig:cover}), the
first multi-robot spatial perception system
 capable of building a hierarchical 3D scene graph online from sensor data.
The system builds on top of Hydra~\cite{Hughes22rss-hydra} and its main features include:
(i) a 3D-scene-graph-based hierarchical loop closure detection module based on~\cite{Hughes22rss-hydra}, which enables a more versatile inter-robot loop closure detection;
(ii) a frame alignment module that removes the need to calibrate the initial poses of the robots;
(iii) an align-optimize-reconcile framework that uses Graduated Non-Convexity (GNC)~\cite{Yang20ral-GNC} to optimize multi-robot 3D scene graphs while being robust to outlier loop closures and erroneous associations of scene graph nodes across different robots.

We evaluate \hydraMulti both in simulation and with a team of robots mapping two student residences.
Our experiments show that 
(i) we can reconstruct a centralized 3D scene graph of large indoor environments with multiple robots online,
(ii) our multi-robot system achieves performance comparable to the single-robot system~\cite{Hughes22rss-hydra},
while enabling faster mapping,
(iii) our system is robust to perceptual aliasing and is able to accurately find the global frame for all the robots in the team without any initial calibration,
and then correctly reconcile and merge the 3D scene graphs obtained from each robot,
and (iv) our system can effectively merge different map representations produced by a %
team of heterogeneous robots.

\section{Related Work}
\label{sec:related_work}

\myParagraph{Multi-Robot SLAM}
Many multi-robot SLAM systems can be split into a frontend for intra and inter-robot loop closure detection
and a backend which typically performs pose-graph or factor-graph optimization~\cite{Cadena16tro-SLAMsurvey,Ebadi22arxiv-surveySLAMSubt}.
The literature on multi-robot SLAM can be similarly divided into works that focus on the frontend, the backend, or the system as a whole.

In a centralized setup, 
a common way to obtain loop closures is to use visual place recognition methods~\cite{Oliva01ijcv,Ulrich00icra,Lowe99iccv,Bay06eccv,Sivic03iccv,Arandjelovic16cvpr-netvlad}. 
Some recent works also focus on designing a distributed multi-robot frontend to find loop closures via local communication among the robots~\cite{Cieslewski18icra,Tian19arxiv,Giamou18icra,Tian18rss,VanOpdenbosch19ral-collabVisualSlam,Tardioli15iros}.
Centralized backend approaches collect measurements
at a central server, which computes the trajectory estimates for all
robots~\cite{Andersson08icra,Kim10icra,Bailey11icra,Lazaro11icra,Dong15icra}. 
Distributed backend approaches, on the other hand, distribute the pose graph optimization task among the robots~\cite{Choudhary17ijrr-distributedPGO3D,Aragues11icra-distributedLocalization,Tian21tro-distributedPgo,Tian20ral-asynchronous,Fan20iros-majorization}.
A number of recent papers develop complete systems for localization and mapping,
such as~\cite{Cieslewski18icra,Choudhary17ijrr-distributedPGO3D,Wang19arxiv,Chang22ral-LAMP2,Miller2020ral-mineMultiQuadrupeds,Polizzi2022ral-decentralizedThermal,Lajoie23arxiv-SwarmSlam}. 
However, there is only a sparse set of works that aims at enabling multi-robot systems to construct higher-level metric-semantic representations:
\cite{Chang21icra-KimeraMulti,Tian22tro-KimeraMulti} present a distributed multi-robot system that constructs semantically labeled 3D meshes of the environment,
while \cite{Miller2022ral-StrongerTA} demonstrates a system where ground robots localize in a semantic map created in real time by a high-altitude quadrotor
by semantically matching local maps with the overhead map for cross-view localization.

\myParagraph{Metric-semantic and Hierarchical Mapping}
Spatial AI~\cite{Davison18-futuremapping} is a concept proposed to build perception systems that allow a robot to more 
effectively 
interact with its environment. 
This idea of building maps that are more compatible with high-level tasks assigned to a robot, 
along with the new possibilities from deep learning and the maturity of traditional 3D reconstruction and SLAM techniques,
gave rise to a surge in interest towards metric-semantic and hierarchical mapping.
The literature includes approaches to build maps based on objects~\cite{Salas-Moreno13cvpr,Dong17cvpr-XVIO,Mo19iros-orcVIO,Nicholson18ral-quadricSLAM,Bowman17icra,Ok21icra-home},
volumetric models~\cite{McCormac17icra-semanticFusion,Grinvald19ral-voxbloxpp,Narita19iros-metricSemantic},
point clouds~\cite{Behley19iccv-semanticKitti,Tateno15iros-metricSemantic,Lianos18eccv-VSO},
3D meshes~\cite{Rosinol20icra-Kimera,Rosu19ijcv-semanticMesh},
hierarchies~\cite{Kuipers00ai,Kuipers78cs,Chatila85,Thrun02a,Ruiz-Sarmiento17kbs-multiversalMaps,Galindo05iros-multiHierarchicalMaps,Zender08ras-spatialRepresentations},
and combinations thereof~\cite{Li16iros-metricSemantic,McCormac183dv-fusion++,Xu19icra-midFusion,Schmid21arxiv-panoptic}.
More recently,
3D scene graphs have been proposed as expressive hierarchical models of the environment~\cite{Armeni19iccv-3DsceneGraphs,Rosinol21ijrr-Kimera}.
The approaches in~\cite{Armeni19iccv-3DsceneGraphs,Rosinol21ijrr-Kimera,Rosinol20rss-dynamicSceneGraphs} are designed for offline use, while Wu\setal~\cite{Wu21cvpr-SceneGraphFusion}, Hughes\setal~\cite{Hughes22rss-hydra}, and Bavle\setal~\cite{Bavle2022ral-SGraph,Bavle22arxiv-Sgraphs+}
have recently demonstrated online construction of 3D scene graphs.
\section{\hydraMulti}
\label{sec:system}
\begin{figure}[!t]
    \centering
    \includegraphics[trim={0 0 0 0},clip, width=1.0\columnwidth]{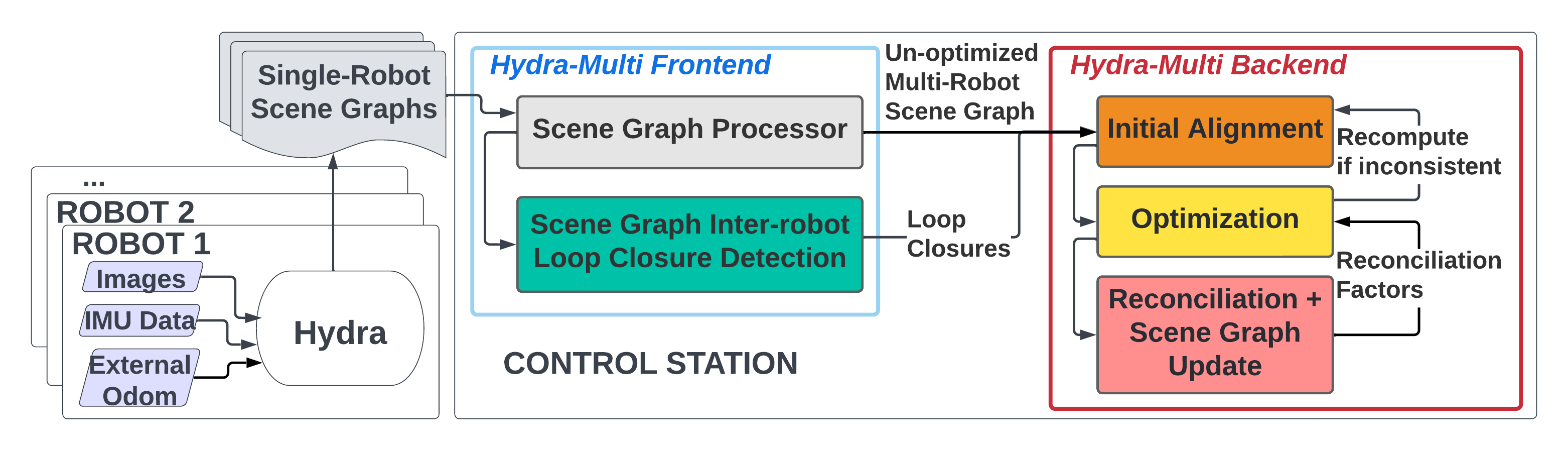}\vspace{-3mm}
    \caption{ The \hydraMulti system consists of a multi-robot frontend and a multi-robot backend. The frontend is in charge of processing the single-robot inputs and detecting inter-robot loop closures. The backend executes our alignment, optimization, and reconciliation framework.}\label{fig:architecture}
\end{figure}

The architecture of the proposed spatial perception system, dubbed \emph{\hydraMulti}, is shown in~\cref{fig:architecture}.
It consists of a frontend for interfacing the centralized control station with the individual robots and for loop closure detection,
and a backend for scene graph optimization and reconciliation.
In this section, we discuss each component of \hydraMulti. %

\subsection{\hydraMulti Frontend}
Each robot runs a local instance of Hydra~\cite{Hughes22rss-hydra} to build a local 3D scene graph of the portion of the environment it has explored.
The \hydraMulti frontend, running at the control station, receives the latest scene graphs from the individual robots, and is also in charge of detecting loop closures between the robots (\ie \emph{inter-robot loop closures}).

\myParagraph{Scene Graph Processor}
The partial single-robot 3D scene graphs are first collected into a single un-optimized and un-reconciled frontend scene graph. The frontend scene graph is \emph{un-optimized} in that it might suffer from large drift since it does not include inter-robot loop closures; it might even include scene graphs that do not share a common reference frame (see discussion about Initial Alignment below).
Moreover, it is \emph{un-reconciled}, since it might have many duplicated nodes (\eg multiple nodes corresponding to the same object observed by different robots).
When the control station receives the latest scene graph from a robot (the entire scene graph is transmitted periodically from each robot to the control station),
new nodes and edges are added,
and the nodes and edges that were deleted on the robot scene graph are also removed from the frontend scene graph.
Careful book-keeping is done on tracking the reference frame of each node for the initial alignment step in the backend.
The frontend scene graph is kept separate from the backend scene graph, which instead is optimized and reconciled at each iteration.

\myParagraph{Loop Closure Detection}
To search for inter-robot loop closures, 
we apply the hierarchical loop closure detection method described in~\cite{Hughes22rss-hydra} to the  multi-robot frontend scene graph.
The method consists of a \emph{Top-down Loop Closure Detection} step that compares scene-graph-based descriptors computed across different layers of the frontend  
scene graph (from places, which are representations of free space in the environment, to objects, to visual appearance descriptors),
and a \emph{Bottom-up Geometric Verification} step that attempts to register the matches at each layer with RANSAC (for visual keypoints) or  TEASER++~\cite{Yang20tro-teaser} (for object nodes).

\subsection{\hydraMulti Backend}\label{sec:backend}

\begin{figure}[!t]
    \centering
    \includegraphics[trim={20 20 20 20},clip, width=1.01\columnwidth]{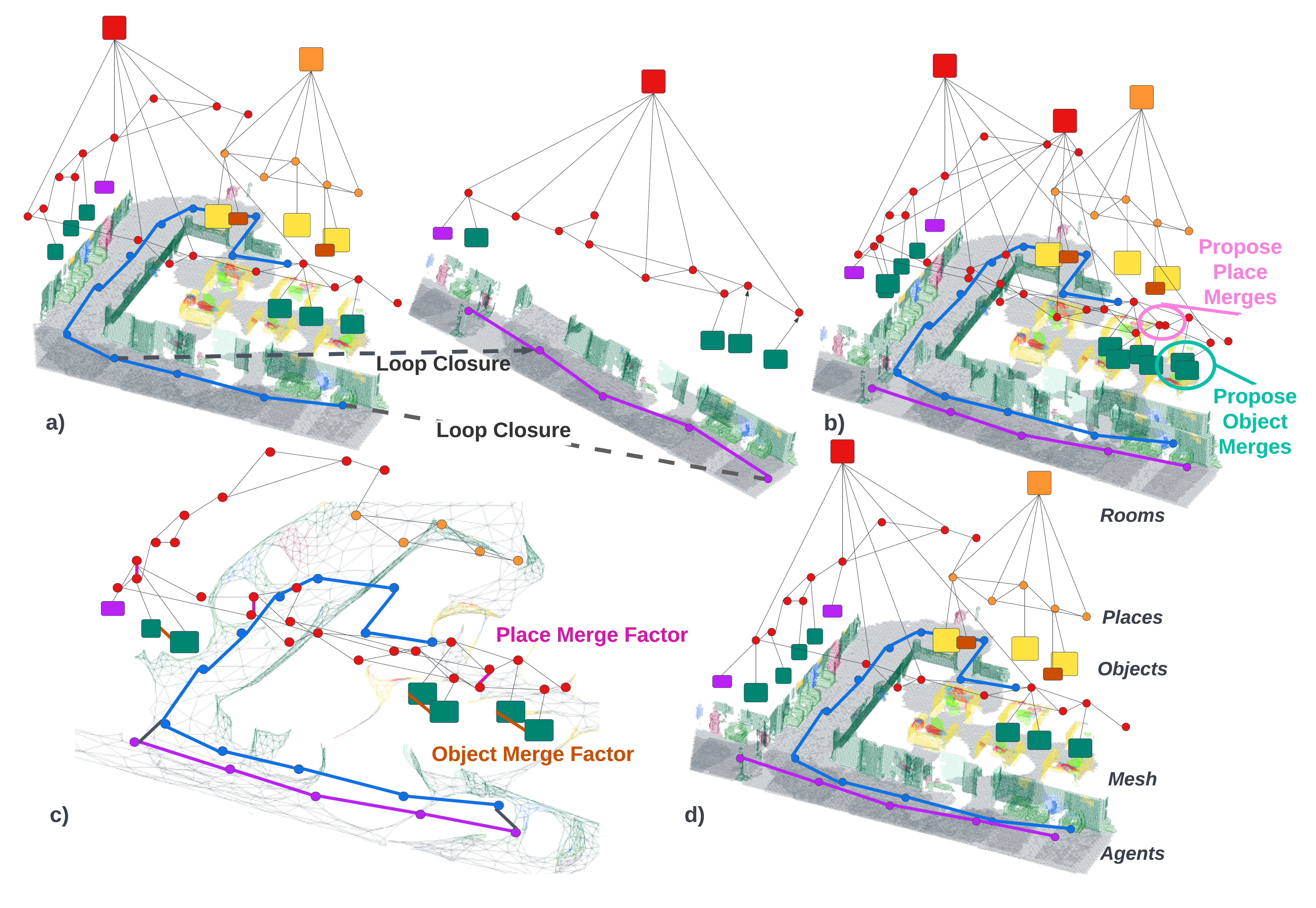}
    \vspace{-8mm}
    \caption{(a) The \hydraMulti frontend detects a loop closure.
    (b) An initial alignment step uses the detected inter-robot loop closures and robust pose-averaging to 
estimate the relative pose between the robots;
    at this stage, candidate node merges are also proposed.
    (c) An optimization of the full scene graph is performed using an embedded deformation graph approach, as in~\cite{Hughes22rss-hydra}; candidate merges are added as robust factors to the optimization and optimized using GNC~\cite{Yang20ral-GNC}.
    (d) Based on the results of the optimization, candidate merges selected as inliers by GNC~\cite{Yang20ral-GNC} are merged; the full scene graph is updated based on the solution of the scene graph optimization.
     \label{fig:backend}}
\end{figure}

The backend is in charge of optimizing the frontend scene graph into a globally consistent and unified representation, and reconciling redundant nodes in regions of the environment observed by multiple robots.
Towards this goal, 
the \hydraMulti backend performs the series of operations shown in~\cref{fig:backend}.
These steps include 
(i) finding an initial frame alignment using the detected inter-robot loop closures,
(ii) proposing node reconciliations based on overlapping nodes after initial alignment,
(iii) performing robust pose-graph optimization using an embedded deformation graph,
and (iv) merging the valid reconciled nodes. %
We explain each step below.

\myParagraph{Initial Alignment}
In this step we estimate a common reference frame for all robots,
which we use to transform all the local 3D scene graphs into a common frame
in preparation for 3D scene graph optimization.
To compute an initial multi-robot alignment,
we follow the approach proposed in~\cite{Tian22tro-KimeraMulti}
by first choosing an arbitrary spanning tree in the robot-level dependence graph~\cite{Tian21tro-distributedPgo},
where the nodes of the graph represent the robots and the edges correspond to the inter-robot loop closures, 
and then estimating the relative pose between pairs of robots corresponding to the edges in the spanning tree.

To estimate the relative pose between the reference frames of two robots, say $A$ and $B$,
we take each inter-robot loop closure $(\alpha_i,\beta_j)$ ---where $\alpha_i$ is the node corresponding to the pose of robot $A$ at time $i$, and $\beta_j$ corresponds to the pose of robot $B$ at time $j$--- and obtain a noisy estimate of the pose of the reference frame of $B$ with respect to $A$ as
\begin{equation}
\widehat{\MX}^{A}_{B,{ij}} =
\widehat{\MX}^A_{\alpha_i}
\widetilde{\MX}^{\alpha_i}_{\beta_j}
\big(\widehat{\MX}^B_{\beta_j} \big)^{-1},
\label{eq:naive_inter_frame}
\end{equation}
where $\widehat{\MX}^A_{\alpha_i}$ and $\widehat{\MX}^B_{\beta_j}$ are the odometric estimates 
of the poses of nodes $\alpha_i$ and $\beta_j$,
and $\widetilde{\MX}^{\alpha_i}_{\beta_j}$ is the loop closure measurement.

After computing the relative transformation between the robots for each inter-robot loop closure,
we obtain a set of noisy relative transformation estimates.
In order to obtain a reliable estimate of the true relative transformation, 
we formulate and solve the following robust pose averaging problem,
\begin{align}
\widehat{\MX}^A_B & \in \argmin_{\MX \in \SEthree}
\;\; \sum_{(i,j) \in L_{A,B}} \rho\left(\left\| 
\MX \boxminus \widehat{\MX}^{A}_{B,{ij}}
\right\|_{\Sigma} \right),
\label{eq:pose_avg}
\end{align}
where $\rho : \mathbb{R}_{\geq0} \to \mathbb{R}_{\geq0}$ is the truncated least squares (TLS) robust cost function~\cite{Yang20ral-GNC}, 
 $L_{A,B}$ is the set of inter-robot loop closures between robots $A$ and $B$, 
 and for two poses $\MX$ and $\MY$, we use the standard notation $\MX \boxminus \MY$ to denote the tangent-space representation of the relative pose $\MX\inv \MY$.
 For a vector $\vxx$, we also use the standard notation $\|\vxx\|^2_{\Sigma} = \vxx\tran \Sigma\inv \vxx$ for the Mahalanobis distance.
  In~\eqref{eq:pose_avg}, 
$\Sigma \in \mathbb{S}_{++}^{6}$ is a fixed covariance matrix
and each residual measures the geodesic distance between the to-be-computed average pose $\MX$ and the measurement $\widehat{\MX}^{A}_{B,{ij}}$.
In practice, we solve~\eqref{eq:pose_avg} using the GNC~\cite{Yang20ral-GNC} implementation available in GTSAM~\citep{gtsam}.
We count a robot as initialized if there are at least $k$ inliers detected by GNC ($k\!=\!5$ in our tests).
By chaining the relative poses obtained from \eqref{eq:pose_avg} along the spanning tree, we obtain transforms from 
each robot's local frame to the global frame (conventionally set to be the reference frame of one of the robots).
We then apply this estimated transform to the nodes of the 3D scene graph of each robot to obtain the initial guess for multi-robot 3D scene graph optimization.
Once the global frame for a robot is computed we do not re-compute the initial alignment after detecting more inter-robot loop closures, unless there is a distinct disagreement with the
result of the multi-robot scene graph optimization, \ie if the distance between the relative translation estimate  from the initial alignment and the corresponding  translation computed from the result of the 3D scene graph optimization (discussed below) is greater than a threshold ($10$m in our tests).

\myParagraph{Reconciliation Proposal}
Based on the results of the initial alignment, we identify merge candidates for pairs of place and object nodes. Place merges are proposed if two place nodes overlap (\eg have distance $\leq 0.01$~m) and have similar sizes (\eg difference between radii $\leq 0.01$~m). Object merges are proposed if two objects have the same semantic label and overlapping bounding boxes. 
We only propose as merge candidates nodes belonging to initialized robots.
The places merge candidates are added to the deformation graph (see below) as relative pose factors with identity transform.
The transform for the object merge candidates is first determined by running ICP on the corresponding mesh vertices, and
then added to the deformation graph for scene graph optimization.

\myParagraph{Robust Scene Graph Optimization}
Given the partial single-robot scene graphs, 
the merge candidates, and the inter-robot loop closures, the backend scene graph is optimized using an
\emph{embedded deformation graph} approach~\cite{Sumner07siggraph-embeddedDeformation}. 
The embedded deformation graph approach associates a local frame (\ie a pose) to a subset of nodes in the scene graph
and then solves an optimization problem to adjust the local frames in a way that minimizes deformations associated to each edge (including loop closures). 
The deformation graph can be seen as the factor graph representation of the 3D scene graph,
and the optimization minimizes edges potentials.

More in detail, from the multi-robot scene graph,
the nodes on the agent layer (corresponding to the robot's trajectory),
the places layer, 
and objects that are proposed as merge candidates are added as nodes to the deformation graph,
while a subset of the vertices of the 3D mesh reconstructed by each robot is added as control point vertices to the deformation graph.
Then, the configuration of these poses is estimated by minimizing a cost that captures (i) odometry and loop closure measurements, (ii) local rigidity of the 3D meshes, (iii) reconciliation factors corresponding to merge candidates between nodes of different robots. %
As shown in~\cite{Rosinol21ijrr-Kimera}, assuming rigid transformations for each frame, 
 deformation graph optimization can be formulated as a pose-graph optimization problem: %
\begin{equation}
\label{eq:pgo}
  \argmin_{\substack{\MT_1,\ldots,\MT_{n} \in \SEthree}} \sum_{\ME_{ij}} ||\MT_i^{-1}\MT_j \boxminus \ME_{ij}||^2_{\M\Omega_{ij}}
\end{equation}
where $\MT_i$ is the to-be-estimated pose of each frame $i$ in the deformation graph, and 
$\ME_{ij}$ are the edges in the deformation graph, which correspond to odometric measurements, loop closures, and merge candidates.
We solve~\eqref{eq:pgo} using GNC~\cite{Yang20ral-GNC} in GTSAM~\cite{Dellaert12tr},
in order to reject spurious loop closures and incorrect merge candidates as outliers.

\myParagraph{Node Reconciliation}
Once the optimization is finished, the place nodes are updated with their new positions and the full mesh is interpolated from the new control vertices~\cite{Sumner07siggraph-embeddedDeformation}. 
We then recompute the object centroids and bounding boxes from the position of the corresponding vertices in the newly deformed mesh~\cite{Hughes22rss-hydra}.
A merge candidate is considered valid if it is selected as an inlier by GNC.
After the valid merges are identified, we compare the number of valid merges with the number of proposed merges,
and we undo all merges in the scene graph if the ratio is below a certain threshold ($0.5$ in our experiments).
Finally, we segment rooms from the merged places using the room segmentation approach described in~\cite{Hughes22rss-hydra}.

\subsection{\hydraMulti with Heterogeneous Teams}

\begin{figure}[!t]
    \centering
    \includegraphics[trim={0 0 0 0},clip, width=0.99\columnwidth]{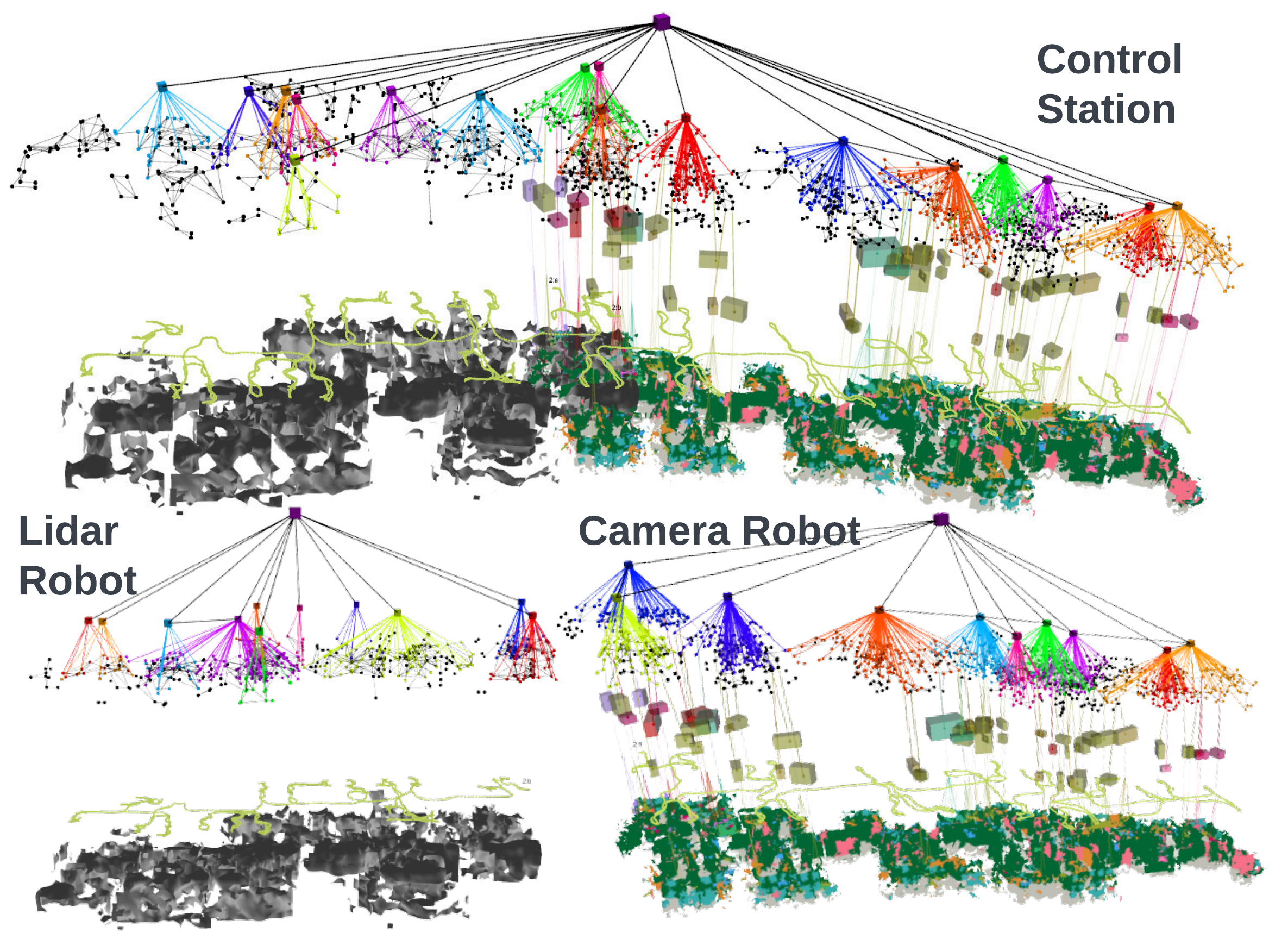}
    \caption{\hydraMulti fusing maps from heterogeneous robots: one robot equipped with visual-inertial sensors produces a semantically annotated map (right-hand side of the scene graph), the other equipped with LIDAR produces a purely geometric 3D reconstruction (left-hand side). 
    \hydraMulti is able to combine both into a unified multi-robot scene graph (top). 
    }\label{fig:hetero_exp}
\end{figure}

The assumed default configuration for each robot in the \hydraMulti system relies on visual-inertial odometry as the localization backbone, and uses RGB-D data for 3D mapping. 
In case a robot with a different sensor or mapping suite is added to the team,
we are still able to support the robot and merge its local map into the \hydraMulti scene graph,  as long as the robot map representation is compatible with at least one layer in the scene graph.
For instance, if a robot performs object-based SLAM with a stereo camera, we can fuse the objects it detects into the object layer of the scene graph. Similarly, if a robot is equipped with a LIDAR and builds a mesh without semantic annotations, we can fuse the mesh and the corresponding graph of places in the scene graph. In other words, the multi-layered nature of the 3D scene graph enables the fusion of heterogeneous maps (\cref{fig:hetero_exp}).

\section{Experiments}
\label{sec:experiments}

This section demonstrates that \hydraMulti builds accurate 3D scene graphs online using inputs from multiple (possibly heterogeneous) robots, %
and provides an ablation of the contribution of each module
 to the performance of the system. %

\subsection{Experimental Setup}

\myParagraph{Datasets}
We use three datasets for our experiments: uHumans2 (uH2), SidPac (SP), and Simmons (SM).
The uH2 dataset~\cite{Rosinol21ijrr-Kimera} is a Unity-based simulated dataset that provides visual-inertial data, as well as ground-truth depth and 2D semantic segmentation.
In particular, we test \hydraMulti on three sequences from a single-floor office environment.

The SP dataset is a real-world dataset collected in a graduate student housing building using a visual-inertial hand-held device as described in~\cite{Hughes22rss-hydra}.
We test \hydraMulti by treating two separate recordings as two different robots.
These scenes are particularly challenging given their scale (average traversal of around 400 meters),
with one recording covering floor 1 and floor 3 of the building
and the other covering floor 3 and floor 4 of the building;
moreover, the robots start on different floors, providing a unique challenge in terms of finding the correct relative transforms between the two robots.
The proxy ground-truth trajectory for this dataset is generated via hand-tuned pose-graph optimization as described in~\cite{Hughes22rss-hydra}.

The SM dataset is a real world dataset collected in an undergraduate student housing building using two Clearpath Jackal robots
equipped with an Intel Realsense D455 camera and a LIDAR.
This dataset is challenging due to its  scale (average traversal of around 500 meters),
and the presence of perceptual aliasing in loop closure detection, caused by visual and structural similarity across different rooms 
(\eg student rooms have similar layout and identical furniture), which leads 
 the frontend to detect up to $80\%$ to $90\%$ outlier loop closures.
Additionally, the robots traverse cluttered rooms and tight spaces, which provide a challenge for \hydra and \hydraMulti.
The proxy ground-truth robot trajectories for this dataset are obtained 
by running a LIDAR SLAM pipeline with flat ground assumption~\cite{Reinke22ral-LOCUS2,Chang22ral-LAMP2}.

\myParagraph{\hydraMulti System}
For all the datasets, we use the RGB-D version of Kimera-VIO~\cite{Rosinol21ijrr-Kimera} for visual-inertial odometry.
For SP, we fuse the Kimera-VIO estimates with the RealSense T265 camera's builtin visual odometry to improve the quality of the trajectory estimate.
However, this is still our most challenging dataset due to unfavorable lighting, prevalence of glass 
(causing
partial and noisy depth estimates from the RGB-D camera), feature-poor regions in hallways, 
and the lack of other types of sensors to improve the odometry estimate.
For SM, we fuse the Kimera-VIO estimates with the Jackal wheel odometry and LIDAR odometry~\cite{Reinke22ral-LOCUS2}, to improve the odometry in the presence of reflective surfaces and large windows.

For SP, we use HRNet~\cite{Wang21pami-hrnet} for 2D semantic segmentation while for SM, we use OneFormer~\cite{Jain22arxiv-oneformer},
which provides a cleaner segmentation from the low viewpoint of the Jackal robots moving in cluttered  environments.

The \hydraMulti system described in Section~\ref{sec:system} is implemented in C++ and ROS.
The experiments are performed on a workstation with a 12-core Intel i9 processor and 2 Nvidia Titan RTX GPUs.

\subsection{Results}

\myParagraph{Localization Error}
We show that \hydraMulti achieves a localization accuracy comparable to state-of-the-art vision-based 
and LIDAR-based SLAM pipelines, 
by comparing it against the system presented in~\cite{Chang22ral-LAMP2} (LAMP 2.0) using the recorded LIDAR and wheel odometry as input, 
and against a centralized version of the system presented in~\cite{Tian22tro-KimeraMulti} (Kimera-Multi) using the same sensor configurations as the \hydraMulti system.
The results are shown in Table~\ref{tab:ate_comparison}.
\hydraMulti leads to slightly better errors compared to the Kimera-Multi pipeline, thanks to the 
hierarchical loop closure detection and the places and object reconciliation.
For the datasets where robots were equipped with LIDAR sensors (SM1, SM2), the performance of 
\hydraMulti remains close to the performance achieved by the LIDAR-based pipeline.

\begin{table}[t!]
\setlength{\tabcolsep}{2pt}
\centering
\caption{ATE (meters) compared to other multi-robot systems.}\label{tab:ate_comparison}
\begin{tabular}{c cccc}
\toprule
& uH2 & SP & SM1 & SM2 \\
\midrule
\multirow{1}{*}{Kimera-Multi~\cite{Tian22tro-KimeraMulti}}
& 0.59 $\pm$ 0.01& 4.99 $\pm$ 1.42& 2.0 $\pm$ 0.18& 0.87 $\pm$ 0.22\\
\multirow{1}{*}{LAMP 2.0~\cite{Chang22ral-LAMP2}}
& - & - & \textbf{0.73 $\pm$ 0.4} & \textbf{0.58 $\pm$ 0.03}\\
\multirow{1}{*}{Hydra-Multi}
& \textbf{0.25 $\pm$ 0.05} & \textbf{3.92 $\pm$ 0.9} & 0.99 $\pm$ 0.31& 0.79 $\pm$ 0.15\\
\midrule
\end{tabular}

\vspace{-1mm}
\end{table}
 
\myParagraph{Objects, Places, and Rooms}
To evaluate \hydraMulti{}'s ability to estimate higher-level abstractions, we first construct a ground-truth scene graph for each dataset.
First, the ground-truth or proxy ground-truth trajectories for each dataset are rigidly aligned to the first robot's reference frame via a hand-tuned ICP registration of the robots' 3D meshes.
Then, we use the aligned ground-truth trajectories for each dataset to reconstruct an %
Euclidean Signed Distance Function (ESDF) and a Generalized Voronoi Diagram (GVD), which we use to examine the accuracy of the places layer,
and a 3D metric-semantic mesh, which we use to extract object locations and bounding boxes.
Ground-truth room bounding-boxes are hand-labeled using the 3D metric semantic mesh for each dataset.

We consider the same performance metrics  used in~\cite{Hughes22rss-hydra}.
For the objects, we consider two metrics:
\percCorrect{} is the percentage of object nodes in the estimated scene graph  that are within a given distance threshold from the corresponding ground-truth object,
and \percFound{} is the percentage of ground-truth objects that are within a given distance threshold\footnote{$1$m for uH2, $3$m for SP, and $2$m for SM based on size of scene and ATE.} from the object nodes in the estimated scene graph. %
For the places, we record the distance of each place node in the scene graph being analyzed to the nearest voxel in the GVD, which we refer to as the \positionError{}.
Finally, we report two metrics for the rooms using the free-space voxels contained within each ground-truth or estimated room.
The first is \roomPrec: the maximum number of overlapping voxels between any ground-truth room and the estimated room node being analyzed.
The second is \roomRec: the maximum number of overlapping voxels between any estimated room and the ground-truth room being analyzed.
These are averaged across all rooms being analyzed.

We compare \hydraMulti against two baselines:
(i) \emph{GT align}, obtained by combining the single-robot scene graphs after aligning them to the ground truth trajectory: this represents the results we would obtain if we did not fuse the results in a centralized map, but rather had each robot explore in isolation after carefully calibrating their initial reference frames;
(ii) %
 \emph{HM align}, obtained by combining the single-robot scene graphs after transformation to the global reference frame using the initial alignment from \hydraMulti: this represents the results we would obtain when each robot maps on its own, but the robots are not initially calibrated.

\begin{figure}[!t]
    \centering
    \includegraphics[trim={0 0 0 0},clip, width=0.99\columnwidth]{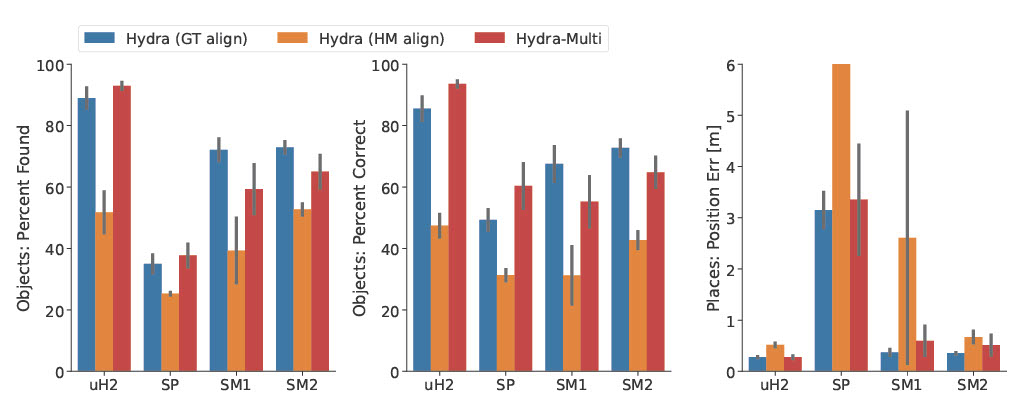}
    \vspace{-8mm}
    \caption{Accuracy of the objects and places estimated by \hydraMulti against two baselines (GT align and HM align). %
     Each plot reports the mean across 5 trials along with the standard deviation as an error bar.}\label{fig:objects_accuracy}
\end{figure}

\begin{figure}[!t]
    \centering
    \includegraphics[trim={0 0 0 0},clip, width=0.99\columnwidth]{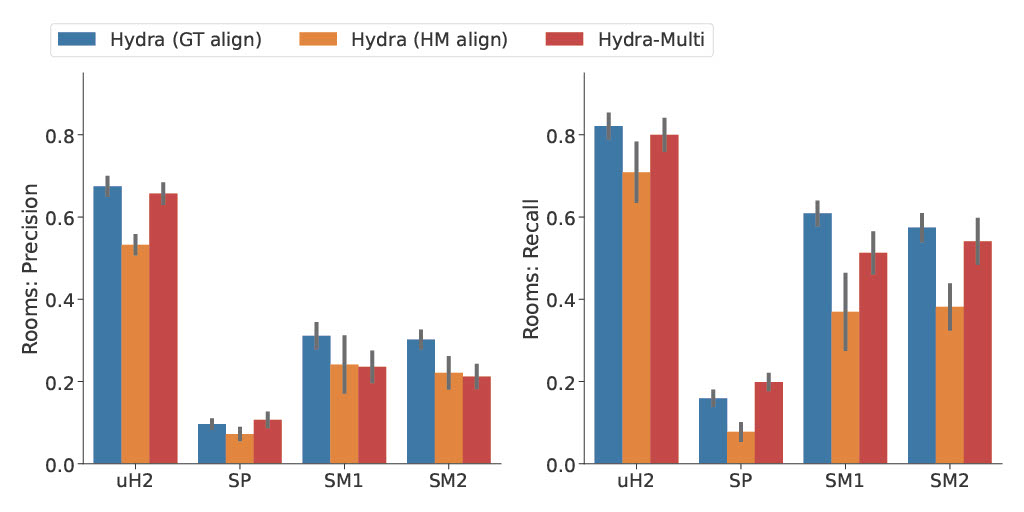}
        \vspace{-8mm}
    \caption{Room detection accuracy estimated by \hydraMulti against two baselines (GT align and HM align).
     Each plot reports the mean across 5 trials along with the standard deviation as an error bar.}\label{fig:rooms_accuracy}
\end{figure}

We run five trials of \hydraMulti on each dataset and 
show the resulting objects and places metrics in Fig.~\ref{fig:objects_accuracy} and the room metrics in Fig.~\ref{fig:rooms_accuracy}.
In all datasets, we show that the performance of \hydraMulti is comparable with that of single-robot \hydra with ground-truth alignment (GT align)
despite not having any initial knowledge of the global reference frame;
it is also consistently better than single-robot \hydra with the estimated initial alignment (HM align), which underlines the effectiveness of the scene graph optimization.
In uH2 and SP, we even see a slight improvement in performance obtained by \hydraMulti as compared to the \emph{GT align} baseline, which results from the  inter-robot loop closures in \hydraMulti.

\begin{figure}[!t]
    \centering
    \includegraphics[trim={0 0 0 0},clip, width=0.99\columnwidth]{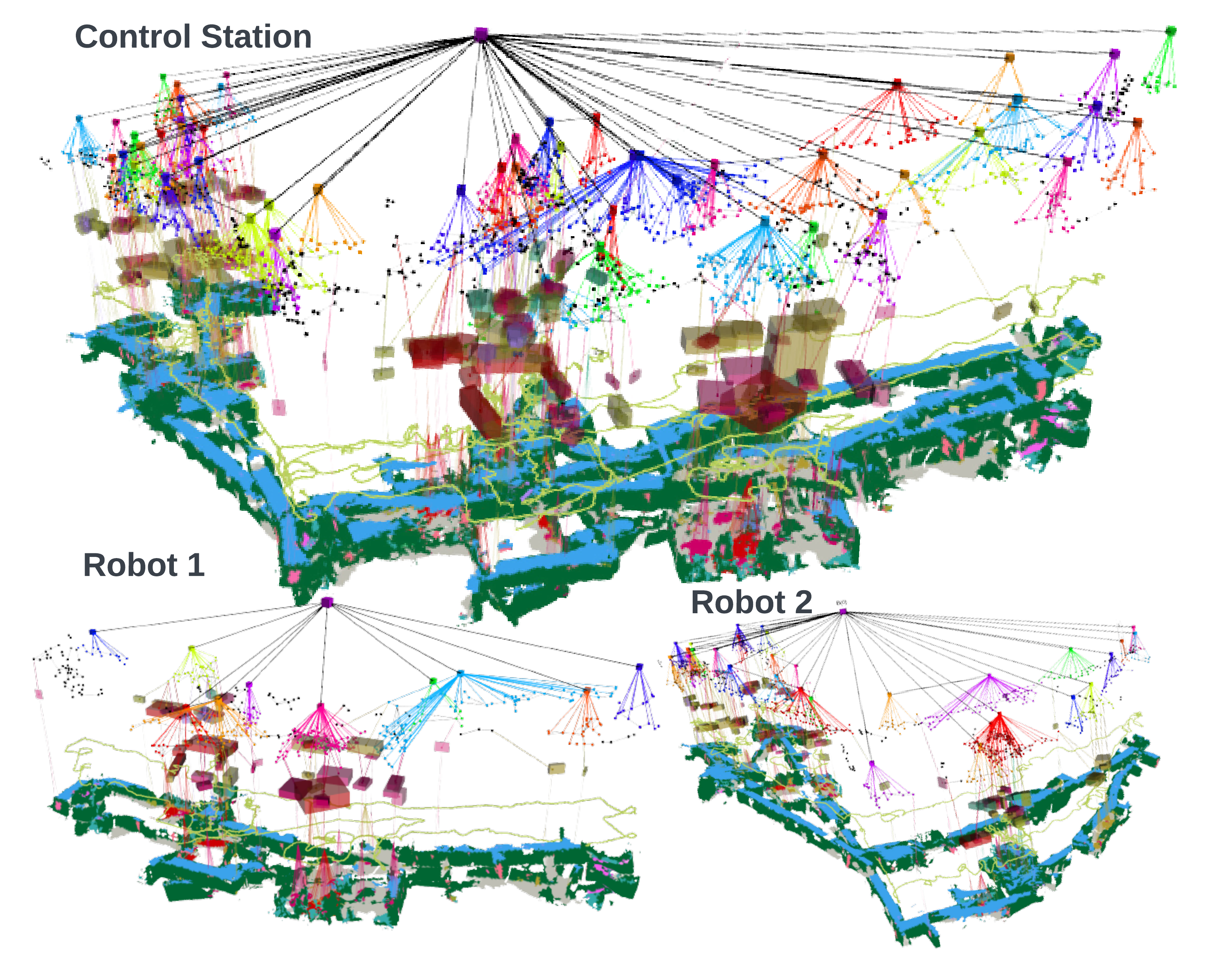}
    \vspace{-10mm}
    \caption{\hydraMulti with two robots exploring the multi-floor environment of the Sidney-Pacific (SP) student residence at MIT.}\label{fig:sidpac_qualitative}
\end{figure}

\begin{figure}[!t]
    \centering
    \vspace{-4mm}
    \includegraphics[trim={0 0 0 0},clip, width=0.99\columnwidth]{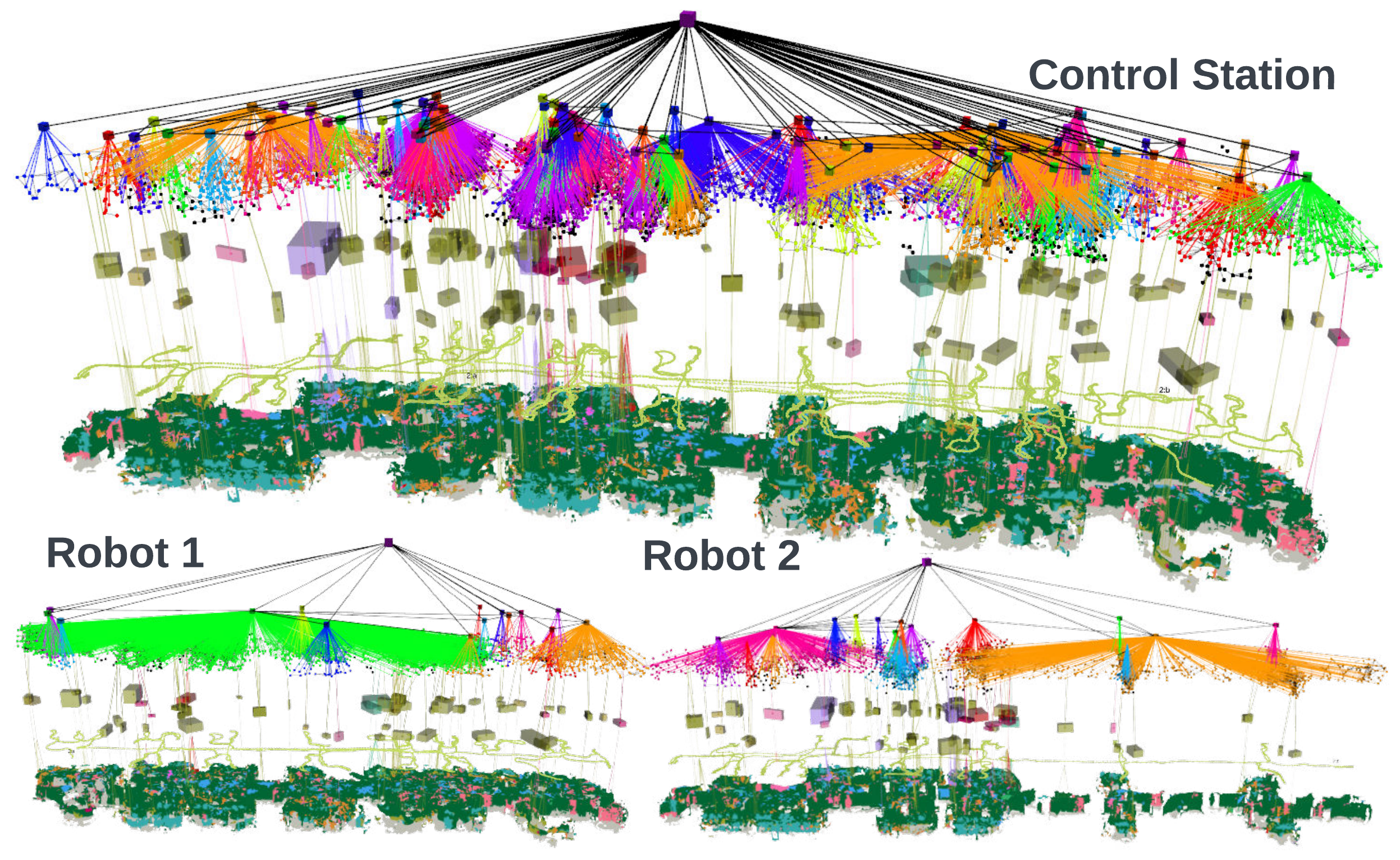}
    \caption{\hydraMulti in the SM1 dataset: two robots start from different rooms and explore different but overlapping areas of the same floor.}\label{fig:simmons_d2r1_qualitative}
\end{figure}

\begin{figure}[!t]
    \centering
    \vspace{-4mm}
    \includegraphics[trim={0 0 0 0},clip, width=0.99\columnwidth]{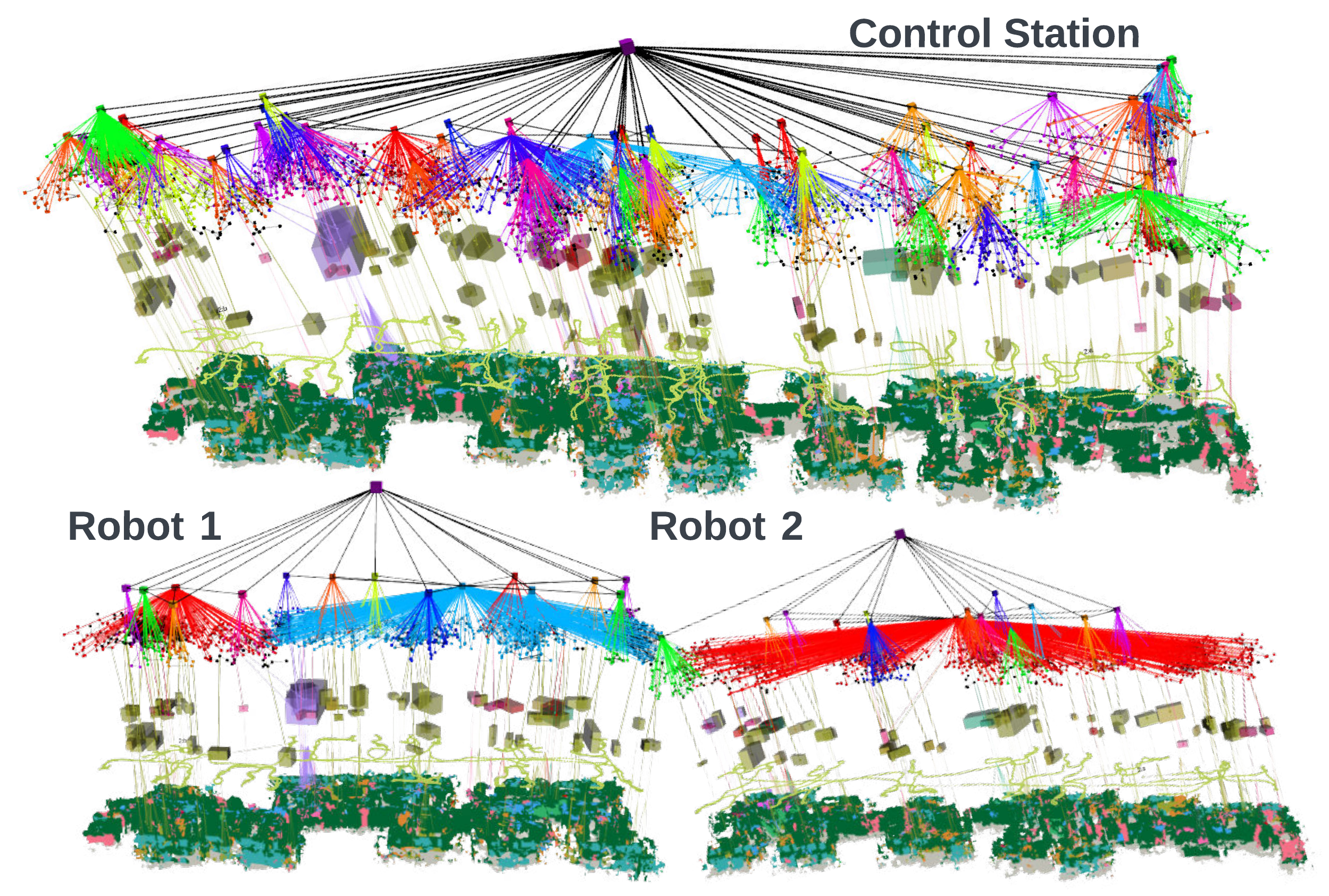}
    \caption{\hydraMulti in the SM2 dataset: two robots start from opposite ends of the same floor and meet in the middle.}\label{fig:simmons_d2r2_qualitative}
\end{figure}

 Qualitative results for the uH2 dataset are shown in Fig.~\ref{fig:cover}, while the SP and SM datasets are shown in Figs.~\ref{fig:sidpac_qualitative}-\ref{fig:simmons_d2r2_qualitative}.
In SM1 (Fig.~\ref{fig:simmons_d2r1_qualitative}), the robots started from different rooms on the same floor of the student dormitory and explored different parts of the scene.
In SM2 (Fig.~\ref{fig:simmons_d2r2_qualitative}), the robots started from different rooms on opposite ends of the floor and rendezvoused in the middle.
In both scenarios, \hydraMulti was successful at reconstructing a 3D scene graph of the complete floor
in around $30$ minutes. Note that covering similar trajectories sequentially using a single robot takes around $50$ minutes, restating the advantage of multi-robot operation.

\begin{figure}[t]
\centering
\subfloat[Runtime of \hydraMulti]{\includegraphics[trim=10 20 90 90, clip, width=0.5\columnwidth]{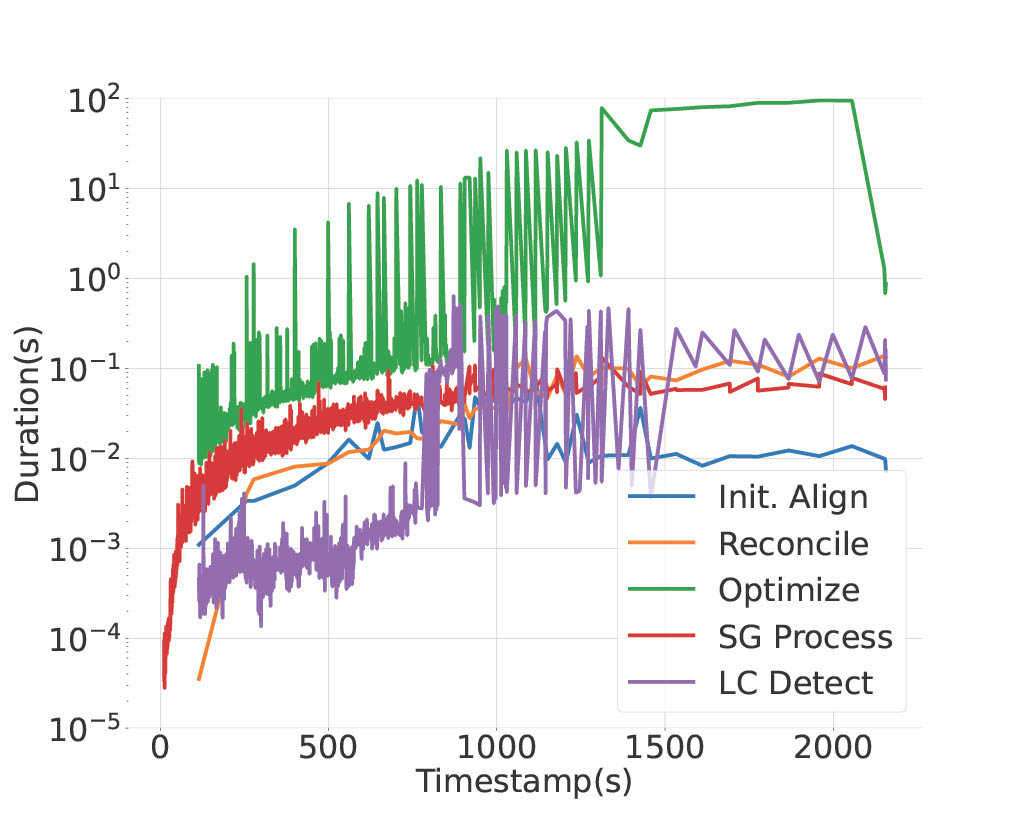}}
\hfill
\subfloat[Bandwidth usage]{\includegraphics[trim=0 0 0 0, clip, width=0.5\columnwidth]{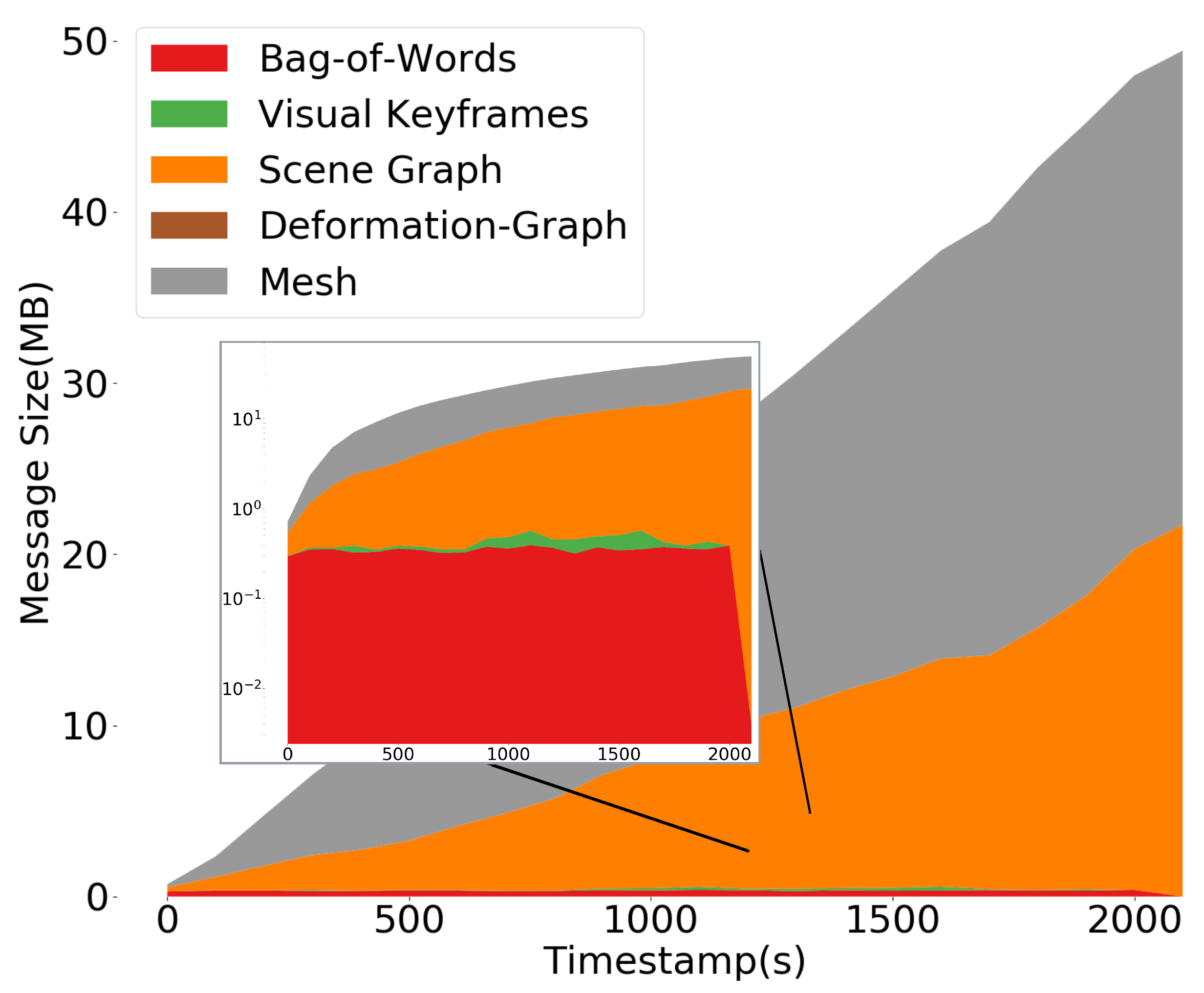}}
\caption{(a) Log plot of the runtime of the \hydraMulti modules over time for the SM2 dataset. (b) Size of the messages transmitted to the control station in the same dataset; the zoomed-in window shows a log plot of the bandwidth usage, to better highlight the contribution of all modules. }
\label{fig:timing_bandwidth}
\end{figure}

\myParagraph{Bandwidth Usage and Runtime}
The runtime of the different components of \hydraMulti for a single run on the SM dataset is shown in Fig.~\ref{fig:timing_bandwidth}(a).
The \hydraMulti frontend time per iteration remains bounded around 100ms, while the backend optimization time increases as the size of the graph grows.
Since the frontend and backend run in parallel threads, 
the \hydraMulti frontend is still able to run in real-time, but the corrections resulting from loop closures must wait for the backend optimization to finish and typically lag behind. %
Fig.~\ref{fig:timing_bandwidth}(b) shows a breakdown of the bandwidth usage.
Most of the bandwidth is consumed by transmitting the mesh and the scene graph,
since the current system naively sends the entire scene graph from each robot to the control station at every update; the bandwidth usage for transmitting additional information for loop closure detection 
(bag-of-word descriptors, features for each keyframe) and inputs to the deformation graph optimization (deformation graph) ---whose computation is entrusted to each robot--- remains below 1MB.

\myParagraph{Components Ablation}
We evaluate the contribution of each component in the backend by running three variants of \hydraMulti{}:
(i) without reconciliation proposals (label ``No Rec''),
(ii) without initial alignment (label ``No IA''),
and (iii) with all proposed features (label ``Full'').
We compare the two object metrics (\percFound{} and \percCorrect) for each variant across 5 trials in Table~\ref{tab:ablation}.
Running with all the components gives the best and most consistent result.
Running without reconciliation produces fewer factors in the optimization to fully align the  scene graphs from the two robot, resulting in a slight accuracy degradation.
Running without initial alignment means that we do not have a good initialization for scene graph optimization, which largely degrades the results.

\begin{table}[t!]
\setlength{\tabcolsep}{2pt}
\centering
\caption{Ablation table of components in the \hydraMulti backend.}\label{tab:ablation}
\begin{tabular}{cl cccc}
\toprule
& & uH2 & SP & SM1 & SM2 \\
\midrule
\multirow{2}{*}{No IA}
& Found (\%) & 80.8 $\pm$ 7.4& 21.4 $\pm$ 12.9& 23.8 $\pm$ 16.6& 27.3 $\pm$ 24.8\\
& Correct (\%) & 90.0 $\pm$ 6.6& 36.8 $\pm$ 20.5& 16.2 $\pm$ 10.1& 26.6 $\pm$ 24.1\\
\midrule
\multirow{2}{*}{No Rec}
& Found (\%) & 91.1 $\pm$ 0.4& 29.1 $\pm$ 13.0& 37.3 $\pm$ 7.1& 54.0 $\pm$ 14.8\\
& Correct (\%) & 92.6 $\pm$ 1.9& 47.7 $\pm$ 22.6& 29.5 $\pm$ 8.9& 52.9 $\pm$ 18.1\\
\midrule
\multirow{2}{*}{Full}
& Found (\%) & \bf{92.9 $\pm$ 1.2}& \bf{37.7 $\pm$ 3.7}& \bf{59.3 $\pm$ 8.0}& \bf{65.0 $\pm$ 5.3}\\
& Correct (\%) & \bf{93.6 $\pm$ 1.0}& \bf{60.3 $\pm$ 7.3}& \bf{55.2 $\pm$ 8.2}& \bf{64.8 $\pm$ 4.9}\\
\midrule
\end{tabular}

\vspace{-1mm}
\end{table}
 
\myParagraph{Heterogeneous Robot Teams}
We perform an experiment on the SM2 dataset with one robot using the default configuration (visual-inertial odometry and RGB-D data as input to the local \hydra),
while another robot uses visual-inertial odometry with LIDAR point clouds as input to the local \hydra.
Fig.~\ref{fig:hetero_exp} shows qualitative results from the \hydraMulti reconstruction.
The LIDAR robot does not perform object detection or semantic segmentation,
hence the gray mesh and the lack of objects in the object layer of the scene graph corresponding to the half of the environment that the LIDAR robot explored;
 however, the robot is still able to contribute to the place layer and mesh of the resulting 
 scene graph.

\section{Conclusion and Future Work} %
\label{sec:conclusions}
We present \hydraMulti,
the first system for collaborative spatial perception
that leverages a control station to receive incremental partial 3D scene graphs from a team of robots
and construct a unified scene graph of the environment.

Future work includes relaxing the assumption that objects in the scene are static, 
and moving from a centralized to a distributed backend for increased scalability.
In particular, we plan to investigate how to reduce the communication bandwidth 
from the agents to the base station 
(as shown in Fig.~\ref{fig:timing_bandwidth}), speed up the scene graph optimization,
and harden \hydraMulti towards large-scale real-world deployment. %

{\footnotesize
\bibliographystyle{IEEEtran}

\begin{thebibliography}{10}
\providecommand{\url}[1]{#1}
\csname url@samestyle\endcsname
\providecommand{\newblock}{\relax}
\providecommand{\bibinfo}[2]{#2}
\providecommand{\BIBentrySTDinterwordspacing}{\spaceskip=0pt\relax}
\providecommand{\BIBentryALTinterwordstretchfactor}{4}
\providecommand{\BIBentryALTinterwordspacing}{\spaceskip=\fontdimen2\font plus
\BIBentryALTinterwordstretchfactor\fontdimen3\font minus
  \fontdimen4\font\relax}
\providecommand{\BIBforeignlanguage}[2]{{%
\expandafter\ifx\csname l@#1\endcsname\relax
\typeout{** WARNING: IEEEtran.bst: No hyphenation pattern has been}%
\typeout{** loaded for the language `#1'. Using the pattern for}%
\typeout{** the default language instead.}%
\else
\language=\csname l@#1\endcsname
\fi
#2}}
\providecommand{\BIBdecl}{\relax}
\BIBdecl

\bibitem{Armeni19iccv-3DsceneGraphs}
I.~Armeni, Z.~He, J.~Gwak, A.~Zamir, M.~Fischer, J.~Malik, and S.~Savarese,
  ``{3D} scene graph: A structure for unified semantics, {3D} space, and
  camera,'' in \emph{Intl. Conf. on Computer Vision (ICCV)}, 2019, pp.
  5664--5673.

\bibitem{Rosinol20rss-dynamicSceneGraphs}
A.~Rosinol, A.~Gupta, M.~Abate, J.~Shi, and L.~Carlone, ``{3D} dynamic scene
  graphs: Actionable spatial perception with places, objects, and humans,'' in
  \emph{Robotics: Science and Systems (RSS)}, 2020,
  \linkToPdf{https://arxiv.org/pdf/2002.06289.pdf},
  \linkToVideo{https://www.youtube.com/watch?v=SWbofjhyPzI&feature=youtu.be}.

\bibitem{Rosinol21ijrr-Kimera}
A.~Rosinol, A.~Violette, M.~Abate, N.~Hughes, Y.~Chang, J.~Shi, A.~Gupta, and
  L.~Carlone, ``Kimera: from {SLAM} to spatial perception with {3D} dynamic
  scene graphs,'' \emph{Intl. J. of Robotics Research}, vol.~40, no. 12--14,
  pp. 1510--1546, 2021, arXiv preprint: 2101.06894,
  \linkToPdf{https://arxiv.org/pdf/2101.06894.pdf}.

\bibitem{Hughes22rss-hydra}
N.~Hughes, Y.~Chang, and L.~Carlone, ``{Hydra:} a real-time spatial perception
  engine for {3D} scene graph construction and optimization,'' in
  \emph{Robotics: Science and Systems (RSS)}, 2022,
  \linkToPdf{https://arxiv.org/pdf/2201.13360.pdf}.

\bibitem{Wu21cvpr-SceneGraphFusion}
S.~Wu, J.~Wald, K.~Tateno, N.~Navab, and F.~Tombari, ``{SceneGraphFusion}:
  Incremental {3D} scene graph prediction from {RGB-D} sequences,'' in
  \emph{IEEE Conf. on Computer Vision and Pattern Recognition (CVPR)}, 2021.

\bibitem{Ravichandran22icra-RLwithSceneGraphs}
Z.~Ravichandran, L.~Peng, N.~Hughes, J.~Griffith, and L.~Carlone,
  ``Hierarchical representations and explicit memory: Learning effective
  navigation policies on {3D} scene graphs using graph neural networks,'' in
  \emph{IEEE Intl. Conf. on Robotics and Automation (ICRA)}, 2022,
  \linkToPdf{https://arxiv.org/pdf/2108.01176.pdf}.

\bibitem{Yang20ral-GNC}
H.~Yang, P.~Antonante, V.~Tzoumas, and L.~Carlone, ``Graduated non-convexity
  for robust spatial perception: From non-minimal solvers to global outlier
  rejection,'' \emph{{IEEE} Robotics and Automation Letters ({RA-L})}, vol.~5,
  no.~2, pp. 1127--1134, 2020, arXiv preprint:1909.08605 (with supplemental
  material), \linkToPdf{https://arxiv.org/pdf/1909.08605.pdf}\award{, ICRA Best
  paper award in Robot Vision}.

\bibitem{Cadena16tro-SLAMsurvey}
C.~Cadena, L.~Carlone, H.~Carrillo, Y.~Latif, D.~Scaramuzza, J.~Neira, I.~Reid,
  and J.~Leonard, ``Past, present, and future of simultaneous localization and
  mapping: Toward the robust-perception age,'' \emph{{IEEE} Trans. Robotics},
  vol.~32, no.~6, pp. 1309--1332, 2016, arxiv preprint: 1606.05830,
  \linkToPdf{https://arxiv.org/abs/1606.05830}.

\bibitem{Ebadi22arxiv-surveySLAMSubt}
K.~Ebadi, L.~Bernreiter, H.~Biggie, G.~Catt, Y.~Chang, A.~Chatterjee,
  C.~Denniston, S.-P. Desch\^{e}nes, K.~Harlow, S.~Khattak, L.~Nogueira,
  M.~Palieri, P.~Petr\'{a}\u{c}ek, P.~Petrl\'{i}k, A.~Reinke,
  V.~Kr\'{a}tk\'{y}, S.~Zhao, A.~Agha-mohammadi, K.~Alexis, C.~Heckman,
  K.~Khosoussi, N.~Kottege, B.~Morrell, M.~Hutter, F.~Pauling, F.~Pomerleau,
  M.~Saska, S.~Scherer, R.~Siegwart, J.~Williams, and L.~Carlone, ``Present and
  future of {SLAM} in extreme underground environments,'' \emph{arXiv preprint:
  2208.01787}, 2022, \linkToPdf{https://arxiv.org/pdf/2208.01787.pdf}.

\bibitem{Oliva01ijcv}
A.~Oliva and A.~Torralba, ``Modeling the shape of the scene: a holistic
  representation of the spatial envelope,'' \emph{Intl. J. of Computer Vision},
  vol.~42, pp. 145--175, 2001.

\bibitem{Ulrich00icra}
I.~Ulrich and I.~Nourbakhsh, ``Appearance-based place recognition for
  topological localization,'' in \emph{IEEE Intl. Conf. on Robotics and
  Automation (ICRA)}, vol.~2, April 2000, pp. 1023 -- 1029.

\bibitem{Lowe99iccv}
D.~Lowe, ``Object recognition from local scale-invariant features,'' in
  \emph{Intl. Conf. on Computer Vision (ICCV)}, 1999, pp. 1150--1157.

\bibitem{Bay06eccv}
H.~Bay, T.~Tuytelaars, and L.~V. Gool, ``{SURF}: speeded up robust features,''
  in \emph{European Conf. on Computer Vision (ECCV)}, 2006.

\bibitem{Sivic03iccv}
J.~Sivic and A.~Zisserman, ``Video google: a text retrieval approach to object
  matching in videos,'' in \emph{Intl. Conf. on Computer Vision (ICCV)}, 2003.

\bibitem{Arandjelovic16cvpr-netvlad}
R.~{Arandjelovic}, P.~{Gronat}, A.~{Torii}, T.~{Pajdla}, and J.~{Sivic},
  ``{NetVLAD}: {CNN} architecture for weakly supervised place recognition,'' in
  \emph{IEEE Conf. on Computer Vision and Pattern Recognition (CVPR)}, 2016,
  pp. 5297--5307.

\bibitem{Cieslewski18icra}
T.~Cieslewski, S.~Choudhary, and D.~Scaramuzza, ``Data-efficient decentralized
  visual {SLAM},'' \emph{IEEE Intl. Conf. on Robotics and Automation (ICRA)},
  2018.

\bibitem{Tian19arxiv}
Y.~Tian, K.~Khosoussi, and J.~P. How, ``A resource-aware approach to
  collaborative loop closure detection with provable performance guarantees,''
  \emph{arXiv preprint arXiv:1907.04904}, 2019.

\bibitem{Giamou18icra}
M.~Giamou, K.~Khosoussi, and J.~P. How, ``Talk resource-efficiently to me:
  Optimal communication planning for distributed loop closure detection,'' in
  \emph{IEEE Intl. Conf. on Robotics and Automation (ICRA)}, 2018, pp. 1--9.

\bibitem{Tian18rss}
Y.~Tian, K.~Khosoussi, M.~Giamou, J.~P. How, and J.~Kelly, ``Near-optimal
  budgeted data exchange for distributed loop closure detection,'' in
  \emph{Robotics: Science and Systems (RSS)}, 2018.

\bibitem{VanOpdenbosch19ral-collabVisualSlam}
D.~Van~Opdenbosch and E.~Steinbach, ``Collaborative visual slam using
  compressed feature exchange,'' \emph{{IEEE} Robotics and Automation Letters},
  vol.~4, no.~1, pp. 57--64, 2019.

\bibitem{Tardioli15iros}
D.~{Tardioli}, E.~{Montijano}, and A.~R. {Mosteo}, ``Visual data association in
  narrow-bandwidth networks,'' in \emph{IEEE/RSJ Intl. Conf. on Intelligent
  Robots and Systems (IROS)}, 2015, pp. 2572--2577.

\bibitem{Andersson08icra}
L.~Andersson and J.~Nygards, ``C-{SAM} : Multi-robot {SLAM} using square root
  information smoothing,'' in \emph{IEEE Intl. Conf. on Robotics and Automation
  (ICRA)}, 2008.

\bibitem{Kim10icra}
B.~Kim, M.~Kaess, L.~Fletcher, J.~Leonard, A.~Bachrach, N.~Roy, and S.~Teller,
  ``Multiple relative pose graphs for robust cooperative mapping,'' in
  \emph{IEEE Intl. Conf. on Robotics and Automation (ICRA)}, Anchorage, Alaska,
  May 2010, pp. 3185--3192.

\bibitem{Bailey11icra}
T.~Bailey, M.~Bryson, H.~Mu, J.~Vial, L.~McCalman, and H.~Durrant-Whyte,
  ``Decentralised cooperative localisation for heterogeneous teams of mobile
  robots,'' in \emph{IEEE Intl. Conf. on Robotics and Automation (ICRA)},
  Shanghai, China, May 2011, pp. 2859--2865.

\bibitem{Lazaro11icra}
M.~Lazaro, L.~Paz, P.~Pinies, J.~Castellanos, and G.~Grisetti, ``Multi-robot
  {SLAM} using condensed measurements,'' in \emph{IEEE Intl. Conf. on Robotics
  and Automation (ICRA)}, 2011, pp. 1069--1076.

\bibitem{Dong15icra}
J.~Dong, E.~Nelson, V.~Indelman, N.~Michael, and F.~Dellaert, ``Distributed
  real-time cooperative localization and mapping using an uncertainty-aware
  expectation maximization approach,'' in \emph{IEEE Intl. Conf. on Robotics
  and Automation (ICRA)}, Seattle, WA, May 2015, pp. 5807--5814.

\bibitem{Choudhary17ijrr-distributedPGO3D}
S.~Choudhary, L.~Carlone, C.~Nieto, J.~Rogers, H.~Christensen, and F.~Dellaert,
  ``Distributed mapping with privacy and communication constraints: Lightweight
  algorithms and object-based models,'' \emph{Intl. J. of Robotics Research},
  2017, arxiv preprint: 1702.03435,
  \linkToPdf{http://arxiv.org/abs/1702.03435}.

\bibitem{Aragues11icra-distributedLocalization}
R.~Aragues, L.~Carlone, G.~Calafiore, and C.~Sagues, ``Multi-agent localization
  from noisy relative pose measurements,'' in \emph{IEEE Intl. Conf. on
  Robotics and Automation (ICRA)}, 2011, pp. 364--369.

\bibitem{Tian21tro-distributedPgo}
Y.~Tian, K.~Khosoussi, D.~M. Rosen, and J.~P. How, ``Distributed certifiably
  correct pose-graph optimization,'' \emph{IEEE Transactions on Robotics},
  vol.~37, no.~6, pp. 2137--2156, 2021.

\bibitem{Tian20ral-asynchronous}
Y.~Tian, A.~Koppel, A.~S. Bedi, and J.~P. How, ``Asynchronous and parallel
  distributed pose graph optimization,'' \emph{IEEE Robotics and Automation
  Letters}, vol.~5, no.~4, pp. 5819--5826, 2020.

\bibitem{Fan20iros-majorization}
T.~Fan and T.~Murphey, ``Majorization minimization methods for distributed pose
  graph optimization with convergence guarantees,'' in \emph{2020 IEEE/RSJ
  International Conference on Intelligent Robots and Systems (IROS)}, 2020, pp.
  5058--5065.

\bibitem{Wang19arxiv}
\BIBentryALTinterwordspacing
W.~Wang, N.~Jadhav, P.~Vohs, N.~Hughes, M.~Mazumder, and S.~Gil, ``Active
  rendezvous for multi-robot pose graph optimization using sensing over
  {Wi-Fi},'' \emph{CoRR}, vol. abs/1907.05538, 2019. [Online]. Available:
  \url{http://arxiv.org/abs/1907.05538}
\BIBentrySTDinterwordspacing

\bibitem{Chang22ral-LAMP2}
Y.~Chang, K.~Ebadi, C.~Denniston, M.~F. Ginting, A.~Rosinol, A.~Reinke,
  M.~Palieri, J.~Shi, C.~A, B.~Morrell, A.~Agha-mohammadi, and L.~Carlone,
  ``{LAMP 2.0}: A robust multi-robot {SLAM} system for operation in challenging
  large-scale underground environments,'' \emph{{IEEE} Robotics and Automation
  Letters ({RA-L})}, vol.~7, no.~4, pp. 9175--9182, 2022,
  \linkToPdf{https://arxiv.org/pdf/2205.13135.pdf}.

\bibitem{Miller2020ral-mineMultiQuadrupeds}
I.~D. Miller, F.~Cladera, A.~Cowley, S.~S. Shivakumar, E.~S. Lee,
  L.~Jarin-Lipschitz, A.~Bhat, N.~Rodrigues, A.~Zhou, A.~Cohen, A.~Kulkarni,
  J.~Laney, C.~J. Taylor, and V.~Kumar, ``Mine tunnel exploration using
  multiple quadrupedal robots,'' \emph{IEEE Robotics and Automation Letters},
  vol.~5, no.~2, pp. 2840--2847, 2020.

\bibitem{Polizzi2022ral-decentralizedThermal}
V.~Polizzi, R.~Hewitt, J.~Hidalgo-Carrió, J.~Delaune, and D.~Scaramuzza,
  ``Data-efficient collaborative decentralized thermal-inertial odometry,''
  \emph{IEEE Robotics and Automation Letters}, vol.~7, no.~4, pp.
  10\,681--10\,688, 2022.

\bibitem{Lajoie23arxiv-SwarmSlam}
P.-Y. Lajoie and G.~Beltrame, ``Swarm-slam : Sparse decentralized collaborative
  simultaneous localization and mapping framework for multi-robot systems,''
  \emph{arXiv:2301.06230 [cs]}, Jan 2023.

\bibitem{Chang21icra-KimeraMulti}
Y.~Chang, Y.~Tian, J.~How, and L.~Carlone, ``{Kimera-Multi}: a system for
  distributed multi-robot metric-semantic simultaneous localization and
  mapping,'' in \emph{IEEE Intl. Conf. on Robotics and Automation (ICRA)},
  2021, arXiv preprint: 2011.04087,
  \linkToPdf{https://arxiv.org/pdf/2011.04087.pdf}.

\bibitem{Tian22tro-KimeraMulti}
Y.~Tian, Y.~Chang, F.~H. Arias, C.~Nieto-Granda, J.~How, and L.~Carlone,
  ``{Kimera-Multi}: Robust, distributed, dense metric-semantic slam for
  multi-robot systems,'' \emph{{IEEE} Trans. Robotics}, 2022, accepted, arXiv
  preprint: 2106.14386, \linkToPdf{https://arxiv.org/pdf/2106.14386.pdf}.

\bibitem{Miller2022ral-StrongerTA}
I.~D. Miller, F.~C. Ojeda, T.~Smith, C.~J. Taylor, and V.~R. Kumar, ``Stronger
  together: Air-ground robotic collaboration using semantics,'' \emph{IEEE
  Robotics and Automation Letters}, vol.~7, pp. 9643--9650, 2022.

\bibitem{Davison18-futuremapping}
A.~J. Davison, ``{FutureMapping}: The computational structure of spatial {AI}
  systems,'' \emph{arXiv preprint arXiv:1803.11288}, 2018.

\bibitem{Salas-Moreno13cvpr}
R.~F. Salas-Moreno, R.~A. Newcombe, H.~Strasdat, P.~H.~J. Kelly, and A.~J.
  Davison, ``{SLAM++}: Simultaneous localisation and mapping at the level of
  objects,'' in \emph{IEEE Conf. on Computer Vision and Pattern Recognition
  (CVPR)}, 2013.

\bibitem{Dong17cvpr-XVIO}
J.~Dong, X.~Fei, and S.~Soatto, ``{Visual-Inertial-Semantic} scene
  representation for {3D} object detection,'' in \emph{IEEE Conf. on Computer
  Vision and Pattern Recognition (CVPR)}, 2017.

\bibitem{Mo19iros-orcVIO}
M.~Shan, Q.~Feng, and N.~Atanasov, ``Object residual constrained
  visual-inertial odometry,'' in \emph{IEEE/RSJ Intl. Conf. on Intelligent
  Robots and Systems (IROS)}, 2020, pp. 5104--5111.

\bibitem{Nicholson18ral-quadricSLAM}
L.~Nicholson, M.~Milford, and N.~S{\"u}nderhauf, ``{QuadricSLAM}: Dual quadrics
  from object detections as landmarks in object-oriented {SLAM},'' \emph{{IEEE}
  Robotics and Automation Letters}, vol.~4, pp. 1--8, 2018.

\bibitem{Bowman17icra}
S.~Bowman, N.~Atanasov, K.~Daniilidis, and G.~Pappas, ``Probabilistic data
  association for semantic {SLAM},'' in \emph{IEEE Intl. Conf. on Robotics and
  Automation (ICRA)}, 2017, pp. 1722--1729.

\bibitem{Ok21icra-home}
K.~Ok, K.~Liu, and N.~Roy, ``Hierarchical object map estimation for efficient
  and robust navigation,'' in \emph{2021 IEEE International Conference on
  Robotics and Automation (ICRA)}, 2021, pp. 1132--1139.

\bibitem{McCormac17icra-semanticFusion}
J.~McCormac, A.~Handa, A.~J. Davison, and S.~Leutenegger, ``{SemanticFusion:
  Dense 3D Semantic Mapping with Convolutional Neural Networks},'' in
  \emph{IEEE Intl. Conf. on Robotics and Automation (ICRA)}, 2017.

\bibitem{Grinvald19ral-voxbloxpp}
M.~{Grinvald}, F.~{Furrer}, T.~{Novkovic}, J.~J. {Chung}, C.~{Cadena},
  R.~{Siegwart}, and J.~{Nieto}, ``{Volumetric Instance-Aware Semantic Mapping
  and 3D Object Discovery},'' \emph{{IEEE} Robotics and Automation Letters},
  vol.~4, no.~3, pp. 3037--3044, 2019.

\bibitem{Narita19iros-metricSemantic}
G.~Narita, T.~Seno, T.~Ishikawa, and Y.~Kaji, ``Panopticfusion: Online
  volumetric semantic mapping at the level of stuff and things,'' in
  \emph{IEEE/RSJ Intl. Conf. on Intelligent Robots and Systems (IROS)}, 2019.

\bibitem{Behley19iccv-semanticKitti}
J.~Behley, M.~Garbade, A.~Milioto, J.~Quenzel, S.~Behnke, C.~Stachniss, and
  J.~Gall, ``{SemanticKITTI: A Dataset for Semantic Scene Understanding of
  LiDAR Sequences},'' in \emph{Intl. Conf. on Computer Vision (ICCV)}, 2019.

\bibitem{Tateno15iros-metricSemantic}
K.~{Tateno}, F.~{Tombari}, and N.~{Navab}, ``Real-time and scalable incremental
  segmentation on dense {SLAM},'' in \emph{IEEE/RSJ Intl. Conf. on Intelligent
  Robots and Systems (IROS)}, 2015, pp. 4465--4472.

\bibitem{Lianos18eccv-VSO}
K.~Lianos, J.~Sch{\"o}nberger, M.~Pollefeys, and T.~Sattler, ``Vso: Visual
  semantic odometry,'' in \emph{European Conf. on Computer Vision (ECCV)},
  2018, pp. 246--263.

\bibitem{Rosinol20icra-Kimera}
A.~Rosinol, M.~Abate, Y.~Chang, and L.~Carlone, ``Kimera: an open-source
  library for real-time metric-semantic localization and mapping,'' in
  \emph{IEEE Intl. Conf. on Robotics and Automation (ICRA)}, 2020, arXiv
  preprint: 1910.02490,
  \linkToVideo{https://www.youtube.com/watch?v=-5XxXRABXJs},
  \linkToCode{https://github.com/MIT-SPARK/Kimera},
  \linkToPdf{https://arxiv.org/pdf/1910.02490.pdf}.

\bibitem{Rosu19ijcv-semanticMesh}
R.~Rosu, J.~Quenzel, and S.~Behnke, ``Semi-supervised semantic mapping through
  label propagation with semantic texture meshes,'' \emph{Intl. J. of Computer
  Vision}, 06 2019.

\bibitem{Kuipers00ai}
B.~Kuipers, ``The {S}patial {S}emantic {H}ierarchy,'' \emph{Artificial
  {I}ntelligence}, vol. 119, pp. 191--233, 2000.

\bibitem{Kuipers78cs}
------, ``Modeling spatial knowledge,'' \emph{Cognitive Science}, vol.~2, pp.
  129--153, 1978.

\bibitem{Chatila85}
R.~Chatila and J.-P. Laumond, ``Position referencing and consistent world
  modeling for mobile robots,'' in \emph{IEEE Intl. Conf. on Robotics and
  Automation (ICRA)}, 1985, pp. 138--145.

\bibitem{Thrun02a}
S.~Thrun, ``Robotic mapping: a survey,'' in \emph{Exploring artificial
  intelligence in the new millennium}.\hskip 1em plus 0.5em minus 0.4em\relax
  Morgan Kaufmann, Inc., 2003, pp. 1--35.

\bibitem{Ruiz-Sarmiento17kbs-multiversalMaps}
J.-R. Ruiz-Sarmiento, C.~Galindo, and J.~Gonzalez-Jimenez, ``Building
  multiversal semantic maps for mobile robot operation,'' \emph{Knowledge-Based
  Systems}, vol. 119, pp. 257--272, 2017.

\bibitem{Galindo05iros-multiHierarchicalMaps}
C.~Galindo, A.~Saffiotti, S.~Coradeschi, P.~Buschka, J.~Fern\'andez-Madrigal,
  and J.~Gonz\'alez, ``Multi-hierarchical semantic maps for mobile robotics,''
  in \emph{IEEE/RSJ Intl. Conf. on Intelligent Robots and Systems (IROS)},
  2005, pp. 3492--3497.

\bibitem{Zender08ras-spatialRepresentations}
H.~Zender, O.~M. Mozos, P.~Jensfelt, G.-J. Kruijff, and W.~Burgard,
  ``Conceptual spatial representations for indoor mobile robots,''
  \emph{Robotics and Autonomous Systems}, vol.~56, no.~6, pp. 493--502, 2008,
  from Sensors to Human Spatial Concepts.

\bibitem{Li16iros-metricSemantic}
C.~{Li}, H.~{Xiao}, K.~{Tateno}, F.~{Tombari}, N.~{Navab}, and G.~D. {Hager},
  ``Incremental scene understanding on dense {SLAM},'' in \emph{IEEE/RSJ Intl.
  Conf. on Intelligent Robots and Systems (IROS)}, 2016, pp. 574--581.

\bibitem{McCormac183dv-fusion++}
J.~McCormac, R.~Clark, M.~Bloesch, A.~Davison, and S.~Leutenegger, ``Fusion++:
  Volumetric object-level {SLAM},'' in \emph{Intl. Conf. on 3D Vision (3DV)},
  2018, pp. 32--41.

\bibitem{Xu19icra-midFusion}
B.~Xu, W.~Li, D.~Tzoumanikas, M.~Bloesch, A.~Davison, and S.~Leutenegger,
  ``{MID-Fusion}: Octree-based object-level multi-instance dynamic {SLAM},''
  2019, pp. 5231--5237.

\bibitem{Schmid21arxiv-panoptic}
L.~Schmid, J.~Delmerico, J.~Sch{\"o}nberger, J.~Nieto, M.~Pollefeys,
  R.~Siegwart, and C.~Cadena, ``Panoptic multi-tsdfs: a flexible representation
  for online multi-resolution volumetric mapping and long-term dynamic scene
  consistency,'' \emph{arXiv preprint arXiv:2109.10165}, 2021.

\bibitem{Bavle2022ral-SGraph}
H.~Bavle, J.~L. Sanchez-Lopez, M.~Shaheer, J.~Civera, and H.~Voos,
  ``Situational graphs for robot navigation in structured indoor
  environments,'' \emph{{IEEE} Robotics and Automation Letters}, vol.~7, no.~4,
  pp. 9107--9114, 2022.

\bibitem{Bavle22arxiv-Sgraphs+}
H.~Bavle, J.~Sanchez-Lopez, M.~Shaheer, J.~Civera, and H.~Voos, ``{S-Graphs+}:
  Real-time localization and mapping leveraging hierarchical representations,''
  \emph{arXiv:2212.11770 [cs]}, Dec 2022.

\bibitem{Yang20tro-teaser}
H.~Yang, J.~Shi, and L.~Carlone, ``{TEASER: Fast and Certifiable Point Cloud
  Registration},'' \emph{{IEEE} Trans. Robotics}, vol.~37, no.~2, pp. 314--333,
  2020, extended arXiv version 2001.07715
  \linkToPdf{https://arxiv.org/pdf/2001.07715.pdf}.

\bibitem{gtsam}
{F. Dellaert et al.}, ``{Georgia Tech Smoothing And Mapping (GTSAM)},''
  \url{https://gtsam.org/}, 2019.

\bibitem{Sumner07siggraph-embeddedDeformation}
R.~Sumner, J.~Schmid, and M.~Pauly, ``Embedded deformation for shape
  manipulation,'' \emph{ACM SIGGRAPH 2007 papers on - SIGGRAPH '07}, 2007.

\bibitem{Dellaert12tr}
F.~Dellaert, ``Factor graphs and {GTSAM}: A hands-on introduction,'' Georgia
  Institute of Technology, Tech. Rep. GT-RIM-CP\&R-2012-002, September 2012.

\bibitem{Reinke22ral-LOCUS2}
A.~Reinke, M.~Palieri, B.~Morrell, Y.~Chang, K.~Ebadi, L.~Carlone, and
  A.~Agha-mohammadi, ``{LOCUS 2.0}: Robust and computationally efficient lidar
  odometry for real-time underground {3D} mapping,'' vol.~7, no.~4, pp.
  9043--9050, 2022, \linkToPdf{https://arxiv.org/pdf/2205.11784.pdf}.

\bibitem{Wang21pami-hrnet}
J.~Wang, K.~Sun, T.~Cheng, B.~Jiang, C.~Deng, Y.~Zhao, D.~Liu, Y.~Mu, M.~Tan,
  X.~Wang, W.~Liu, and B.~Xiao, ``Deep high-resolution representation learning
  for visual recognition,'' \emph{IEEE Transactions on Pattern Analysis and
  Machine Intelligence}, vol.~43, no.~10, pp. 3349--3364, 2021.

\bibitem{Jain22arxiv-oneformer}
J.~Jain, J.~Li, M.~Chiu, A.~Hassani, N.~Orlov, and H.~Shi, ``Oneformer: One
  transformer to rule universal image segmentation,'' \emph{arXiv:2211.06220
  [cs]}, 2022.

\end{thebibliography}

}

\end{document}